\documentclass{article}

\usepackage{xcolor}

\usepackage[eandd, final]{neurips_2026}

\usepackage[utf8]{inputenc} %
\usepackage[T1]{fontenc}    %
\usepackage{hyperref}       %
\usepackage{url}            %
\usepackage{booktabs}       %
\usepackage{amsfonts}       %
\usepackage{nicefrac}       %
\usepackage{microtype}      %
\usepackage{xcolor}         %

\usepackage{graphicx}
\usepackage{subcaption}
\usepackage{enumitem}
\usepackage{xcolor}
\usepackage{amssymb}
\usepackage{pifont}

\usepackage{xspace}
\newcommand{\benchmark}{{\textsc{TerminalWorld}}\xspace}

\usepackage{tikz}

\newcommand{\graycircle}[1]{%
  \tikz[baseline=(char.base)]{
    \node[
      shape=circle,
      fill=gray!30,
      text=black,
      inner sep=0.2pt,
      minimum size=1em,
      font=\sffamily\small
    ] (char) {#1};}}

\newcommand{\parabf}[1]{\noindent\textbf{#1}\xspace}

\usepackage{balance}

\usepackage[capitalize,noabbrev,nameinlink]{cleveref}
\crefname{appendix}{Appendix}{Appendices}

\definecolor{lightcyan}{RGB}{10,110,150}
\hypersetup{
  colorlinks=true,
  linkcolor=lightcyan,
  citecolor=lightcyan,
  filecolor=lightcyan,
  urlcolor=lightcyan
}

\usepackage{wrapfig}

\usepackage{makecell}
\usepackage{multirow}

\title{\benchmark: Benchmarking Agents \\ on Real-World Terminal Tasks}

\author{
  Zhaoyang Chu$^{1}$\quad
  Jiarui Hu$^{2}$\thanks{These authors contributed equally as co-second authors.}\quad
  Xingyu Jiang$^{2}$\footnotemark[1]\quad
  Pengyu Zou$^{2}$\footnotemark[1]\quad
  Han Li$^{2}$\quad
  Chao Peng$^{3}$\quad \\
  \textbf{Peter O'Hearn}$^{1}$\quad
  \textbf{Earl T. Barr}$^{1}$\quad
  \textbf{Mark Harman}$^{1}$\quad
  \textbf{Federica Sarro}$^{1}$\quad
  \textbf{He Ye}$^{1}$\thanks{Corresponding author: \texttt{he.ye@ucl.ac.uk}}\quad \\
  $^{1}$University College London \quad
  $^{2}$Nanjing University \quad
  $^{3}$Tencent \\
  \url{https://terminalworld.ai}
}

\begin{document}

\maketitle
\setcounter{footnote}{0}

\begin{abstract}

We introduce \benchmark, a scalable data engine that automatically reverse-engineers high-fidelity evaluation tasks from ``in-the-wild'' terminal recordings.
Processing 80,870 terminal recordings, the engine yields a full benchmark of 1,530 validated tasks, spanning 18 real-world categories, ranging from short everyday operations to workflows exceeding 50 steps, and covering 1,280 unique commands.
From these, we curate a \textsc{Verified} subset of 200 diverse, manually reviewed tasks.
Comprehensive benchmarking on \benchmark-\textsc{Verified} across eight frontier models and six agents reveals that current systems still struggle with authentic terminal workflows, achieving a maximum pass rate of only 62.5\%.
Moreover, \benchmark captures real-world terminal capabilities distinct from existing expert-curated benchmarks (\emph{e.g.}, Terminal-Bench), with only a weak correlation to their scores (Pearson $r=0.20$).
The automated engine makes \benchmark authentic and scalable \textit{by construction}, enabling it to evaluate agents in real-world terminal environments as developer practices evolve.

\end{abstract}

\section{Introduction}
\label{sec:intro}

Terminal environments serve as a primary action space for autonomous agents to complete diverse tasks in complex software systems.
Powered by advances in \textit{Large Language Models} (LLMs) for multi-step reasoning and tool use, these agents are increasingly capable of automating terminal workflows by issuing commands, composing tools, and interpreting feedback within interactive CLI sessions, exemplified by open-source frameworks (\emph{e.g.}, SWE-agent~\citep{yang2024swe_agent}, OpenHands~\citep{wang2025openhands}) and commercial CLI assistants (\emph{e.g.}, Claude Code~\citep{claude_code}, Codex CLI~\citep{codex_cli}, Gemini CLI~\citep{gemini_cli}).
Yet, how to reliably evaluate these agents on \textit{real-world} terminal tasks remains an open question.

The prevailing answer has been manually curated benchmarks, such as Terminal-Bench~\citep{merrill2026terminalbench} and LongCLI-Bench~\citep{feng2026longcli_bench}, where domain experts author tasks paired with executable environments.
However, experts often tend to prioritize adversarial puzzles to artificially maximize difficulty, thereby diverging from authentic, real-world terminal workflows.
Moreover, this labor-intensive process struggles to scale with evolving terminal practices and diverse emerging tools, leaving benchmarks narrowly scoped and quickly outdated.
While recent automated synthesis methods~\citep{zhu2026termigen, lin2026cli_gym, gandhi2026endless, pi2026nemotron_terminal, wu2026terminaltraj} attempt to bypass this scalability bottleneck, they are primarily designed for training and rarely undergo the rigorous validation required to guarantee true authenticity.
Therefore, we argue that an \textbf{authentic, scalable} evaluation system should be established to answer the key question:
\textit{``How well do terminal agents perform on the real-world tasks that evolve alongside everyday practices?''}

To answer this question, we argue that naturally occurring terminal operations, if faithfully recorded, can be \emph{reverse-engineered} into evaluation tasks that are authentic by construction.
The \texttt{asciinema} platform\footnote{\url{https://asciinema.org/}} makes this feasible: developers voluntarily share terminal session recordings, each with a structured transcript capturing every command and its corresponding system response.
These recordings form a self-curated, human-vetted, and continuously growing corpus of authentic developer workflows.
To systematically exploit this resource, we introduce \benchmark, a data engine that operationalizes this insight: it automatically turns in-the-wild terminal recordings into executable, rigorously validated evaluation tasks.

Turning raw recordings into evaluation tasks requires addressing three practical challenges, which our data engine systematically resolves:
\graycircle{1}~\textit{\underline{\textbf{Recordings are noisy and lack clear intent.}}}
Recording transcripts often contain typos, retries, and verbose system output, without explicit statements of the developer's goal.
We address this by distilling each transcript into two artifacts via an LLM (\emph{e.g.}, Claude Sonnet 4.6~\citep{claude_sonnet_4_6}): an outcome-oriented instruction that captures the developer's underlying intent and a clean command script as the reference solution.
\graycircle{2}~\textit{\underline{\textbf{Recordings do not capture the underlying execution environments.}}}
The transcript captures commands but not the underlying system state of the author's machine.
We employ an LLM agent (\emph{e.g.}, Claude Code~\citep{claude_code}) to reverse-engineer this environment by inferring actual requirements while eliminating hallucinated dependencies.
In particular, the agent physically builds a Docker image, launches the container, and replays the reference solution, using runtime failures as feedback for targeted repair.
\graycircle{3}~\textit{\underline{\textbf{Recordings lack an explicit test suite.}}}
While recordings naturally capture the execution trajectory, they lack an explicit test suite to automatically judge whether the task goal is achieved.
Relying solely on LLMs to statically generate these tests is inherently vulnerable to false negatives (where correct solutions are rejected due to brittle tests) and false positives (where tasks can be trivially bypassed or solved by flawed workflows).
We resolve this by equipping the agent with a trial-based refinement loop to generate and calibrate the test suite via actual execution feedback within the reproduced Docker container.

Running this engine over 80,870 raw \texttt{asciinema} recordings yields 1,530 validated terminal tasks as the full \benchmark benchmark.
The resulting tasks span 18 real-world terminal categories, range from short everyday operations to workflows exceeding 50 steps, and cover 1,280 unique commands, 91\% of which are absent from Terminal-Bench.
Since the pipeline is fully automated and \texttt{asciinema} keeps accumulating new uploads, \benchmark can be re-run as the platform grows, allowing it to scale with evolving developer practices.
Unlike prior benchmarks, \benchmark is authentic and scalable \textit{by construction}.

From this collection, we curate a \textsc{Verified}\footnote{Here, ``\textsc{Verified}'' denotes manual review of label correctness, not proof of conformance to a formal specification.} subset of 200 tasks, each cross-reviewed by the authors who manually execute the reference solution in the reproduced environment and audit the semantic alignment across all artifacts.
While the full set of 1,530 tasks provides a scalable resource of in-the-wild terminal workflows, this \textsc{Verified} subset serves as a rigorous and challenging testbed for benchmarking frontier models and agents on complex, real-world terminal tasks.

Through comprehensive benchmarking experiments using \benchmark-\textsc{Verified} across eight frontier LLMs (\emph{e.g.}, Claude Opus 4.7~\citep{claude_opus_4_7}, GPT-5.5~\citep{gpt_5_5}, Gemini 3.1 Pro~\citep{gemini_3_1_pro}) and six leading terminal agents (\emph{e.g.}, Claude Code~\citep{claude_code}, Codex CLI~\citep{codex_cli}, Gemini CLI~\citep{gemini_cli}), our analysis reveals several key findings.
\graycircle{1}~Frontier LLMs still struggle with real-world terminal tasks: even the best model solves only 62.5\%, while failures expose an \textit{efficiency paradox}, spending extra compute exploring authentic environments without making progress.
\graycircle{2}~Agent frameworks mainly affect cost-effectiveness rather than the underlying capability ceiling, suggesting that practical agents for real-world terminal environments should reduce exploration friction rather than merely add orchestration complexity.
\graycircle{3}~Terminal-Bench scores are only weakly predictive of agent performance on \benchmark-\textsc{Verified} (Pearson $r=0.20$), suggesting that existing expert-curated challenges do not fully capture the capabilities needed for real-world terminal workflows.
\graycircle{4}~Although \benchmark tasks are grounded in real-world human recordings, agents often solve them through different valid command paths rather than mimicking the original human workflows, as reflected by an overall median command-set overlap of only 21.4\%.

\section{Related Work}

We review prior work on terminal agents and benchmarks for evaluating their capabilities.

\parabf{Terminal Agents.}
Terminal agents are designed to autonomously issue commands, compose diverse tools, and execute multi-step workflows while interpreting execution feedback within interactive CLI environments.
Early frameworks (\emph{e.g.}, SWE-agent~\citep{yang2024swe_agent}, OpenHands~\citep{wang2025openhands}) wrap shell commands behind structured tool APIs, allowing agents to resolve tasks through a constrained action schema.
More recent native CLI assistants (\emph{e.g.}, Claude Code~\citep{claude_code}, Codex CLI~\citep{codex_cli}, Gemini CLI~\citep{gemini_cli}) instead expose the raw shell as the primary interface.
This enables agents to invoke arbitrary commands and directly observe execution feedback, leading to tighter tool integration on real-world tasks.
In parallel, a growing body of work seeks to advance terminal agents by synthesizing large-scale environments and leveraging agent trajectories for training~\citep{zhu2026termigen, lin2026cli_gym, gandhi2026endless, wu2026terminaltraj, pi2026nemotron_terminal}, improving the underlying models to drive stronger agentic performance in CLI environments.
This rapid proliferation of terminal agents calls for high-quality benchmarks that faithfully assess their real-world capabilities.

\parabf{Benchmarks for Terminal Agents.}
Terminal-oriented evaluation has progressed from narrow, single-skill tasks to end-to-end agentic assessment.
Earlier benchmarks target isolated command-line abilities,
such as
translation of natural language into shell commands (\textit{e.g.}, NL2Bash~\citep{lin2018nl2bash}) and single-turn command execution with interactive feedback (\textit{e.g.}, InterCode~\citep{yang2023intercode}).
Recent efforts assess agents inside interactive shell sandboxes, covering complex
tasks that require multi-step reasoning and tool use, such as Terminal-Bench~\citep{merrill2026terminalbench} and LongCLI-Bench~\citep{feng2026longcli_bench}.
Despite this progress, current benchmarks rely on manual curation, which is costly to scale and gravitates toward adversarial puzzles to maximize difficulty.
As a result, tasks often diverge from authentic developer workflows, and high scores may not reliably reflect an agent's competence on the routine terminal tasks encountered by practitioners.
To address this, our work automates benchmark construction by reverse-engineering in-the-wild terminal recordings, making evaluation grounded in real-world authenticity and scalable as developer practices evolve.

\section{\benchmark: Scalable Data Engine for Real-World Terminal Tasks}

\begin{figure*}[!t]
	\centering
	\includegraphics[width=0.96\linewidth]{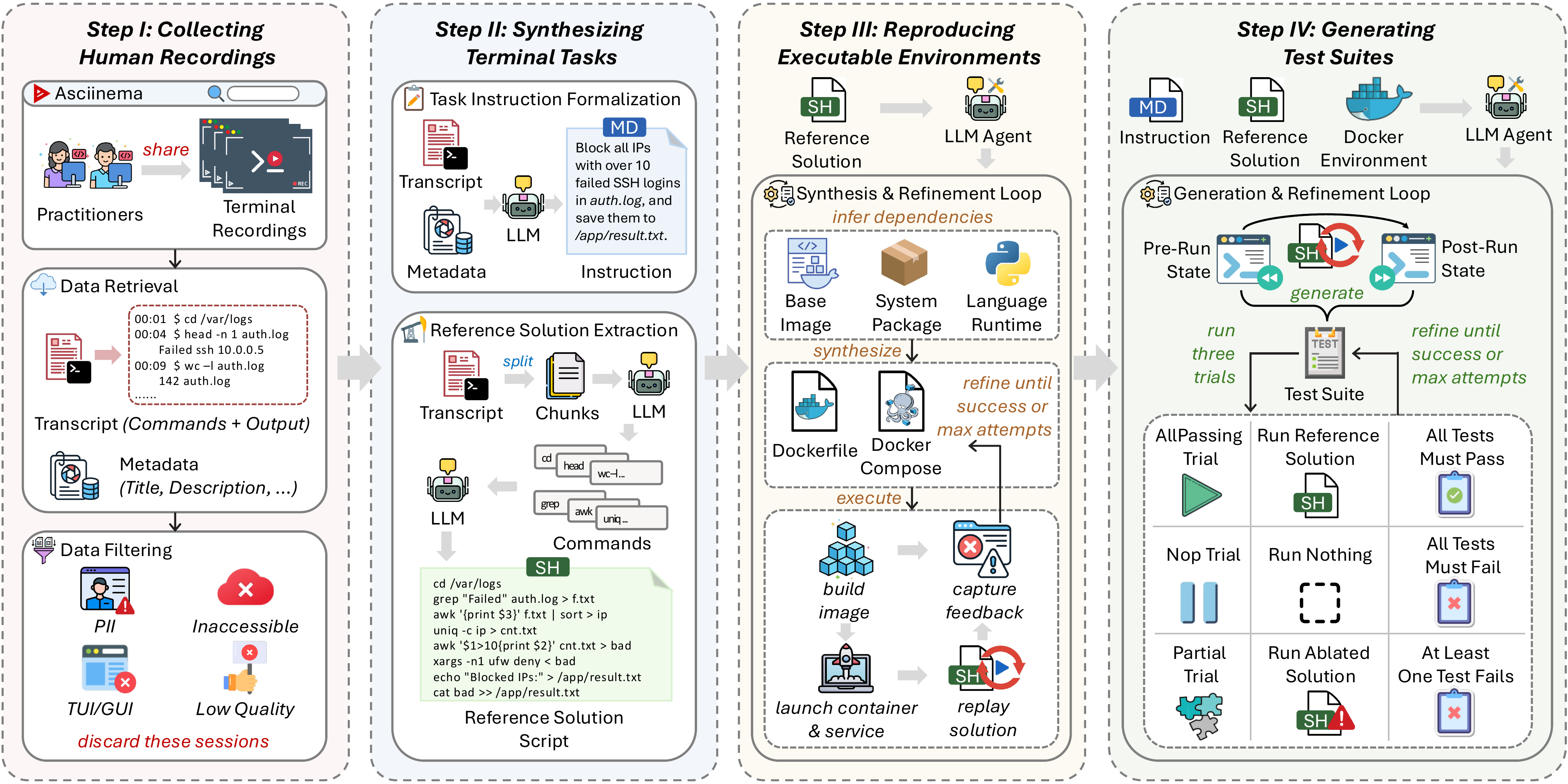}
         \caption{\textbf{An overview of the \benchmark pipeline.} Our data engine automates terminal task synthesis through four key stages: \textbf{(1) Collecting Human Recordings} harvests in-the-wild developer operations; \textbf{(2) Synthesizing Terminal Tasks} distills an outcome-oriented task instruction and a clean reference solution; \textbf{(3) Reproducing Executable Environments} creates and refines an isolated Docker container to replay the core workflow; and \textbf{(4) Generating Test Suites} utilizes a trial-based refinement loop to generate and calibrate test suites within the Docker container.}
	\label{fig_pipeline}
	\vspace{-1em}
\end{figure*}

As illustrated in~\cref{fig_pipeline}, we propose \benchmark, a scalable data engine designed to automatically reverse-engineer terminal tasks\footnote{In this paper, we scope \emph{terminal tasks} to pure CLI workflows, where the agent issues shell commands and observes their \texttt{stdout}/\texttt{stderr} output. TUI-based interactions are outside our current evaluation scope and left to future work; see~\cref{appendix_task_category} for details.} from real-world human recordings, which operates through four key steps:
\textbf{(1) Collecting Human Recordings.}
It harvests large-scale, real-world terminal recordings from the \texttt{asciinema} platform.
\textbf{(2) Synthesizing Terminal Tasks.}
It reverse-engineers the noisy recordings by inferring the underlying human intent to formalize a task instruction and extracting the core commands as the reference solution.
\textbf{(3) Reproducing Executable Environments.}
It creates and refines an isolated Docker container with the corresponding file system and dependencies, ensuring the core recording workflow can be replayed.
\textbf{(4) Generating Test Suites.}
It implements a trial-based refinement loop to generate and calibrate test suites within the reproduced Docker container.

\subsection{Collecting Human Recordings}
\label{sec_data_collection}

We collect real-world terminal operation data from \texttt{asciinema}, a public platform where practitioners share their terminal session recordings.

\parabf{Data Retrieval.}
We systematically index large-scale publicly shared \texttt{asciinema} recordings.
For each recording, we acquire its transcript text via the standard public download links provided by \texttt{asciinema}, along with metadata (\emph{e.g.}, title and description).
The transcript offers a high-fidelity log of the real-world terminal execution, capturing the ordered sequence of executed commands and standard outputs.
In total, we collect 80,870 real-world terminal recordings by humans.

\parabf{Data Filtering.}
While real-world recordings guarantee authenticity, their inherent noise requires systematic filtering.
First, for privacy and safety, we exclude recordings exposing \textit{Personally Identifiable Information} (PII), sensitive credentials, or malicious/destructive commands (\emph{e.g.}, \texttt{rm -rf~*}).
Second, we isolate pure CLI workflows by discarding recordings involving \textit{Text User
Interfaces} (TUIs, \emph{e.g.}, \texttt{vim}, \texttt{nano}, and \texttt{emacs}) or GUI applications.
Third, we remove recordings incapable of deterministic reproduction in Docker containers, such as those dependent on inaccessible URLs, Windows environments, or proprietary software.
Fourth, we eliminate excessively short recordings, typically aborted or trivial sessions.
Finally, we employ an LLM (\emph{e.g.}, Claude Sonnet 4.6~\citep{claude_sonnet_4_6}) to score recording quality, filtering out opaque or purely exploratory sessions (\emph{e.g.}, repetitive \texttt{ls} and \texttt{cat} commands).
Ultimately, this filtering process yields 9,492 high-quality recordings.
Further discussions on ethics, copyright compliance, and privacy mitigation of our data collection are deferred to~\cref{sec:ethics}.

\subsection{Synthesizing Terminal Tasks}
\label{sec_task_synthesis}

Unlike video recordings that require \textit{Vision-Language Models} (VLMs) to parse with substantial overhead and information loss, the \texttt{asciinema} transcript provides a high-fidelity text record of commands and system responses.
We purify the transcript by removing typos, failed attempts, and redundant commands, distilling it into an instruction inferred from the underlying human intent and a clean command sequence as the reference solution.
By reconstructing the developer's goal and workflow, \benchmark ensures that each synthesized task is authentic to real-world usage.

\parabf{Task Instruction Formalization.}
We leverage an LLM (\emph{e.g.}, Claude Sonnet 4.6~\citep{claude_sonnet_4_6}) to synthesize natural language instructions by distilling the developer's core intent from the transcript alongside the title and description.
The instruction specifies the task goal in concise, outcome-oriented language: it must describe the expected \emph{final state}, not the path to reach it.
We explicitly prohibit procedural phrasing, specific commands, and step-by-step enumerations, preventing solution-specific hints from leaking into the task.
To establish a testable contract, the instruction must explicitly specify required output paths (\emph{e.g.}, \texttt{/app/result.txt}) and strict structural formats, while omitting arbitrary internal artifacts invented by the original developer, such as custom labels or print banners.

\parabf{Reference Solution Extraction.}
We employ an LLM (\emph{e.g.}, Claude Sonnet 4.6~\citep{claude_sonnet_4_6}) to extract a clean, executable bash script from the raw transcript as the ground-truth reference solution for the formalized task instruction.
For long transcripts with verbose system outputs, we first split the transcript into chunks to filter execution noise and isolate valid commands.
We then merge the extracted commands, remove duplicates, and assemble a coherent solution workflow.
Since source recordings are mostly pre-planned showcases rather than messy debugging sessions, this process can recover clean reference solutions without being overwhelmed by excessive human trial-and-error.
Consistent with the outcome-oriented instruction design, the script is constrained to redirect its final results from transient terminal outputs to explicit file paths (\textit{e.g.}, \texttt{/app/result.txt}).
This ensures the reference workflow is idempotent and its outcome is deterministically captured in the filesystem.
We generate instructions and reference solutions for all 9,492 filtered recordings.

\subsection{Reproducing Executable Environments}
\label{sec_env_reproduction}

Raw in-the-wild recordings are inherently volatile, as they depend on the original developer's system state, implicit toolchains, and transient resources.
Thus, a basic container for isolated command execution is insufficient; we instead reverse-engineer the dependency context required to replay the recorded terminal workflow.
Without a faithful executable environment, it is difficult to assess the quality of synthesized tasks, such as task solvability and test soundness.
On the other hand, by encapsulating complex dependencies into an isolated Docker sandbox, we eliminate environmental indeterminism for rigorous agent evaluation.

\parabf{Environment Synthesis.}
We leverage an LLM agent (\emph{e.g.}, Claude Code~\citep{claude_code}) to synthesize a \texttt{Dockerfile} (and \texttt{docker-compose.yaml} for multi-service tasks) by inferring required dependencies from the reference solution (\emph{e.g.}, base images, system packages, and language runtimes).
When the recording includes an external repository link, the agent clones and scans the project to infer environment requirements.
To guarantee the environment's authenticity, we eliminate hallucinated dependencies by explicitly prohibiting fake binaries, stubbed dependencies, and bypasses of real software installation.

\parabf{Execution-Based Refinement Loop.}
Since static synthesis by LLMs is prone to dependency conflicts and missing hidden packages, \benchmark equips the agent with an execution-feedback loop to refine the environment.
The agent builds an image from the synthesized Dockerfile, parses build logs to diagnose compilation errors or package manager failures, and iteratively repairs the Dockerfile when needed.
It then runs a Docker container from the image, launching any required auxiliary services via \texttt{docker-compose.yaml} when necessary.
The agent executes the reference solution script in a persistent shell session, aiming to replay the original terminal recording.
The environment is considered reproduced only when the script executes successfully, as indicated by \textit{an exit code of 0}.
Otherwise, any runtime error, such as missing libraries, unconfigured environment variables, or unrecognized commands, is fed back to the agent for targeted repairs to the \texttt{Dockerfile} or \texttt{docker-compose.yaml}.
Recordings that remain irreproducible within our computational budget or are reliant on inaccessible resources are discarded.
Ultimately, this loop reproduces executable environments for 5,035 terminal tasks.

\subsection{Generating Test Suites}
\label{sec_test_generation}

The raw recordings naturally capture the execution trajectory, but they lack explicit tests to automatically judge whether the underlying goal is achieved.
Thus, we introduce an automated execution-feedback loop that generates and refines a test suite within the reproduced Docker environment.

\parabf{Test Suite Generation.}
Based on the formalized instruction (equivalently, the \emph{task specification}) and the reference solution in~\cref{sec_task_synthesis}, test generation reduces to solving the test oracle problem~\citep{barr2015oracle}.
Concretely, we synthesize a suite of test assertions to assess whether the expected final state is achieved.
These assertions typically target persistent artifacts, such as file existence, content hashes, or structured outputs in \texttt{/app/result.txt}.
However, LLM-generated tests without execution feedback are vulnerable to hallucination and misalignment, where ambiguous assertions easily diverge from the actual system state produced by the reference solution.
To resolve this, we adopt an LLM agent (\emph{e.g.}, Claude Code~\citep{claude_code}) to capture snapshots of the pre- and post-execution state in the reproduced Docker environment, recording the true filesystem changes caused by the reference solution.
Using these state deltas, the agent generates and calibrates the test suite to align with the actual final state.
During generation, we explicitly instruct the agent to avoid brittle checks, such as exact string matching on transient outputs or non-deterministic execution values (\textit{e.g.}, timestamps, temporary process IDs).

\parabf{Trial-Based Refinement Loop.}
Once generated, the agent iteratively refines the test suite through three execution trials in fresh, isolated containers to eliminate false negatives and false positives:

\begin{itemize}[leftmargin=*, label=$\triangleright$, itemsep=0.2em, topsep=0.2em]
    \item
    \underline{\textbf{AllPassing Trial}} executes the reference solution and requires all tests to pass.
    It prevents false negatives, where overly rigid tests reject correct solutions, thereby ensuring task \emph{solvability}.
    \item
    \underline{\textbf{Nop Trial}} runs nothing, leaving the container in its initial state, and requires all tests to fail.
    It prevents false positives where tasks are solved by an empty state, thus ensuring task \emph{non-triviality}.
    \item
    \underline{\textbf{Partial Trials}} execute incomplete solutions derived by truncating or ablating the reference solution, and require at least one test to fail.
    This is a stringent check against nuanced false positives, ensuring that the tests can reject incomplete solutions, enhancing the \emph{discriminability} of the test suite.
\end{itemize}

A task is admitted only if its test suite satisfies all three trials.
If any trial fails, the agent diagnoses the trial outputs, applies targeted fixes to the test suite, and reruns the loop.
When necessary, adjustments to the test suite are synchronized back to the instruction, ensuring that the instruction explicitly specifies the state being evaluated.
A task is discarded if its test suite cannot be repaired to pass all three trials within the computational budget.
Ultimately, this refinement loop distills 1,530 validated terminal tasks as the full \benchmark benchmark.
Stage-by-stage filtering and validation statistics are provided in~\cref{appendix_pipeline_statistics}.

\begin{figure*}[!t]
    \centering
    \subcaptionbox{Task Category Distribution.\label{fig_stat_a_task_type}}{
        \includegraphics[width=0.32\linewidth]{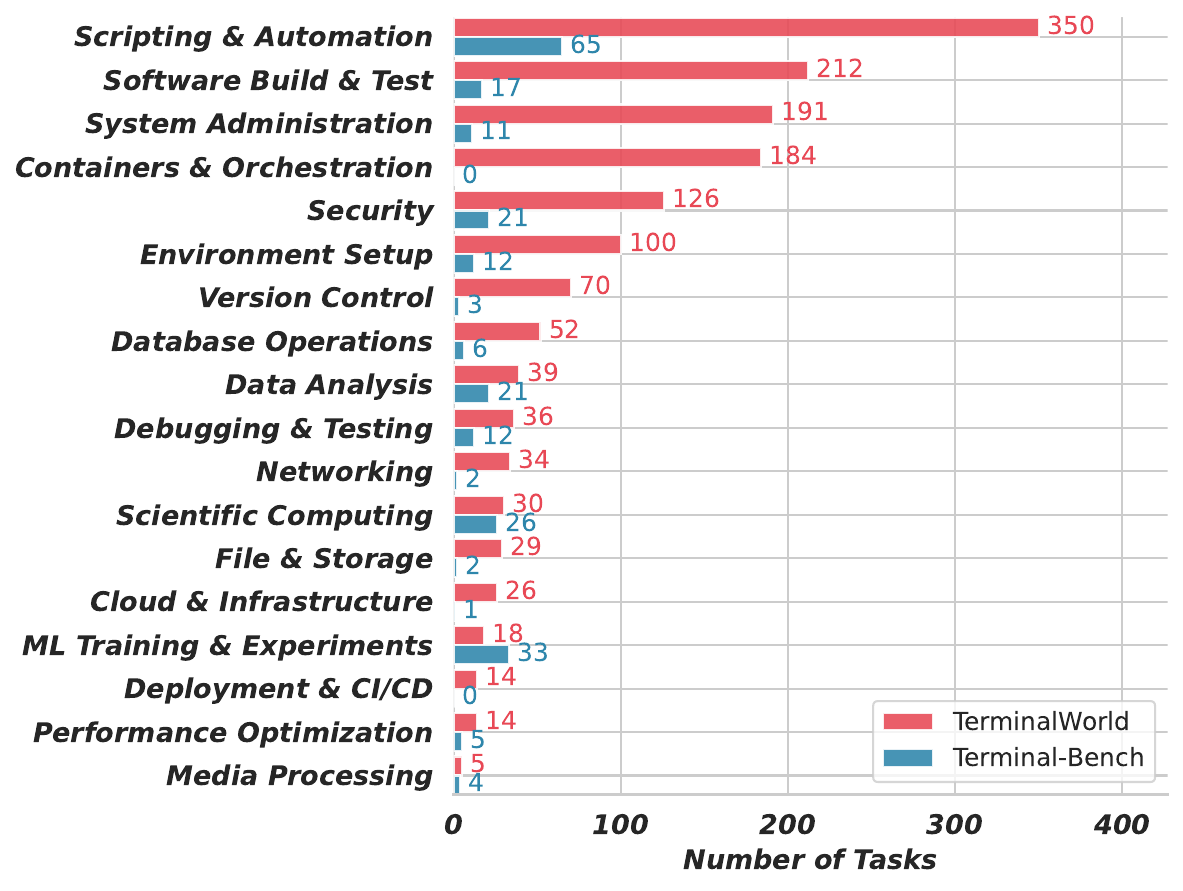}}
    \hfill
    \subcaptionbox{Task Complexity Distribution.\label{fig_stat_b_complexity}}{
        \includegraphics[width=0.32\linewidth]{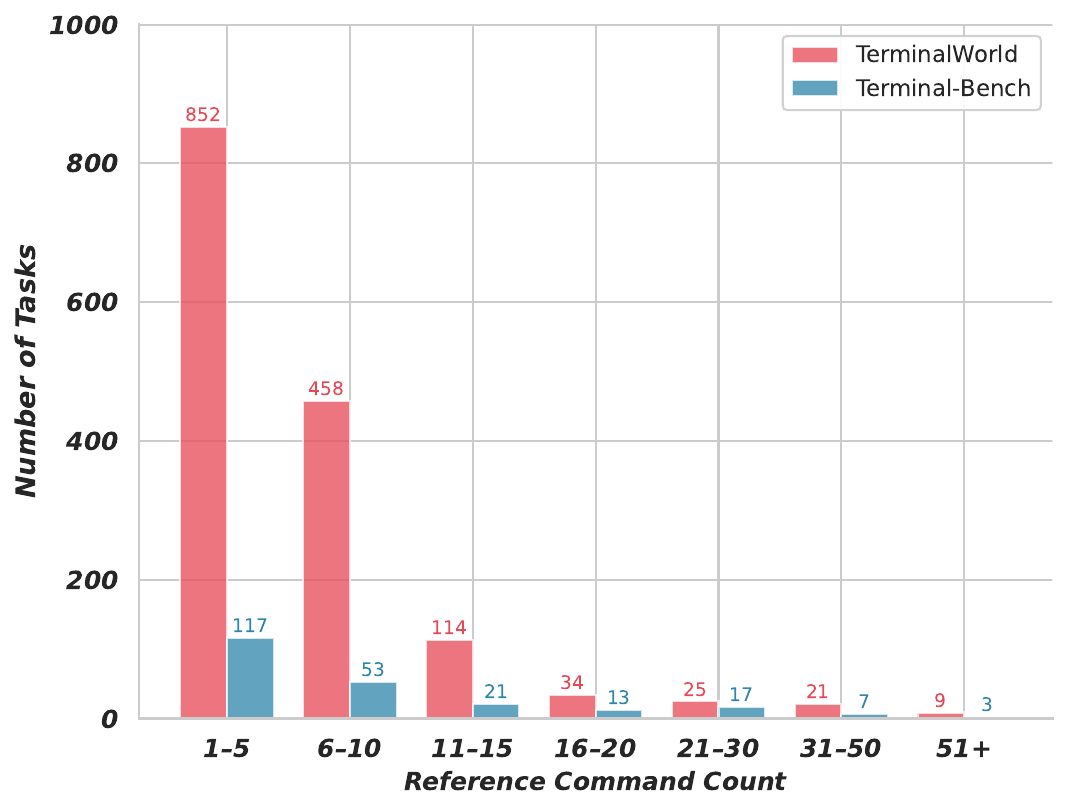}}
    \hfill
    \subcaptionbox{Unique Command Coverage.\label{fig_stat_c_cmd_vocab}}{
        \includegraphics[width=0.32\linewidth]{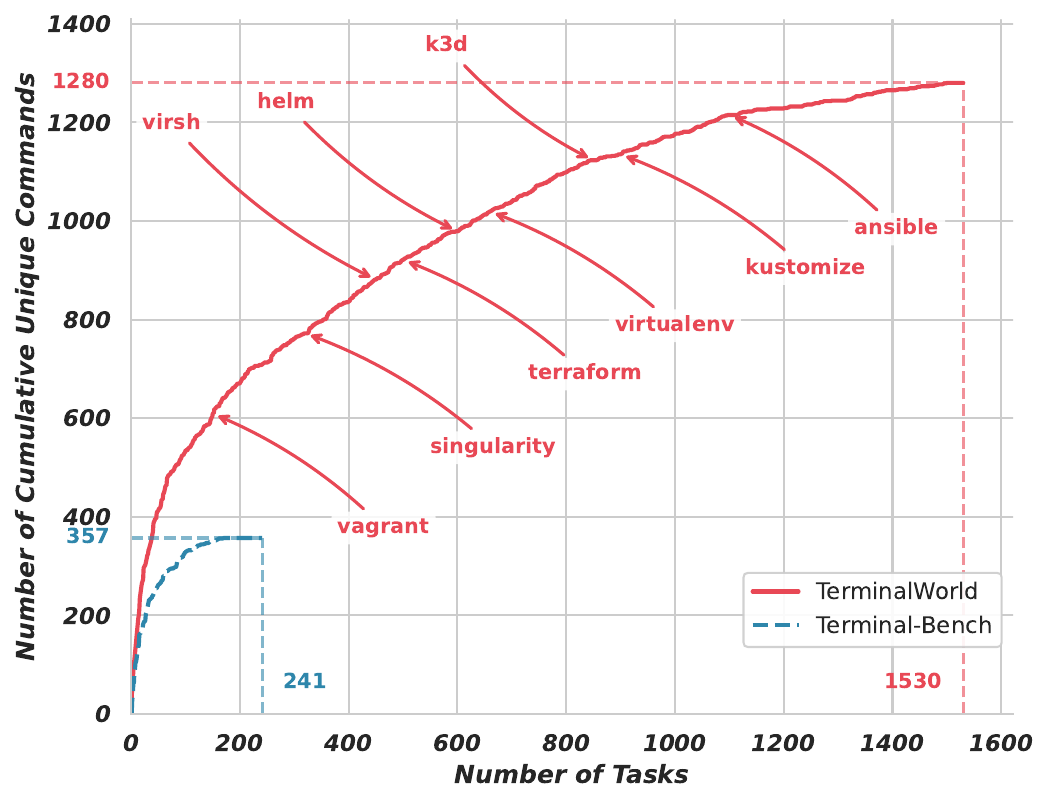}}
    \caption{\textbf{Statistical comparison of 1,530 \benchmark tasks and 241 unique Terminal-Bench tasks.}
    (a) highlights that \benchmark captures diverse real-world workflows (\emph{e.g.}, container orchestration, CI/CD) severely underrepresented in expert-curated benchmarks.
    (b) shows a natural spectrum that mirrors everyday terminal usage, from short operations to multi-step workflows.
    (c) reveals an extensive vocabulary of 1,280 unique commands (91\% absent from Terminal-Bench) that can assess an agent's capability to discover and wield diverse utilities.
    }
    \label{fig_stat_benchmark_comparison}
    \vspace{-1em}
\end{figure*}

\section{The \benchmark Benchmark}

\subsection{Dataset Statistics}
\label{sec_data_stats}

Given that the terminal recordings capture naturally occurring workflows and are voluntarily uploaded by developers, the resulting \benchmark dataset is inherently self-curated and human-vetted.
We obtain both of these highly desirable properties \emph{by construction},
ensuring broad coverage of the tools, configurations, and problem-solving strategies encountered in everyday terminal usage.
We characterize the benchmark's coverage, diversity, and complexity by comparing it against Terminal-Bench~\citep{merrill2026terminalbench}, a widely used benchmark constructed by human experts.

\parabf{Task Category Distribution.}
\cref{fig_stat_a_task_type} details the category distribution of the 1,530 \benchmark tasks alongside the 241 unique tasks de-duplicated across both Terminal-Bench 1.0 and 2.0.
\benchmark covers 18 real-world terminal categories (see~\cref{tab_task_categories} for details), introducing tasks that are ubiquitous in modern developer routines yet severely underrepresented in Terminal-Bench, such as container orchestration, CI/CD pipelines, and cloud infrastructure management.
Beyond the artificial boundaries of expert curation, \benchmark provides a comprehensive and authentic snapshot of the diverse workflows that practitioners actually execute in the wild.

\parabf{Task Complexity Distribution.}
\cref{fig_stat_b_complexity} compares task complexity based on the number of commands required in the reference solutions.
Derived from real-world operations, \benchmark exhibits a natural complexity spectrum spanning from a few commands to workflows exceeding 50 steps, providing a larger scale across all length intervals.
The high density of short workflows reflects everyday terminal usage, where developers frequently execute brief command sequences, while \benchmark still captures the long tail of complex, multi-step scenarios.

\parabf{Unique Command Coverage.}
\cref{fig_stat_c_cmd_vocab} illustrates the cumulative distribution of unique commands appearing in the reference solutions across terminal tasks.
\benchmark's command coverage climbs steadily with task scale, encompassing an extensive vocabulary of 1280 unique commands.
Notably, 91\% of these commands are absent from Terminal-Bench, spanning diverse tool categories such as environment management (\emph{e.g.}, \texttt{vagrant}, \texttt{virtualenv}), infrastructure configuration (\emph{e.g.}, \texttt{terraform}, \texttt{ansible}), and Kubernetes orchestration (\emph{e.g.}, \texttt{k3d}, \texttt{kustomize}).
This broad command coverage reflects the rich diversity of tools developers employ in real-world workflows, establishing \benchmark as an authentic testbed for assessing an agent's capability to discover and wield diverse utilities to solve in-the-wild terminal problems.

\parabf{Intent Fidelity.}
To assess whether the generated tasks preserve the original recording's intent, three authors independently audited 150 tasks uniformly sampled from the full benchmark, examining the generated instructions, reference solutions, and verifiers against the source recordings.
By majority vote, 90.7\% of the tasks were judged faithful, with substantial inter-annotator agreement (Fleiss $\kappa = 0.67$).
The detailed annotation protocol and results are provided in~\cref{appendix_intent_audit}.

\begin{table*}[!t]
\small
\setlength{\tabcolsep}{4.3pt}
\begin{minipage}[t]{\textwidth}
\caption{\textbf{Performance of frontier LLMs on \benchmark under the Terminus-2 scaffold.}
$^\dagger$Values in parentheses denote resolved rate = pass / (total - errors), where errors are tasks on which the Harbor evaluation harness failed before the agent could attempt the task (see~\cref{appendix_llm_benchmarking} for error details).
The results suggest that frontier LLMs still struggle with real-world terminal tasks, often falling into an \textit{efficiency paradox} where more compute and attempts fail to yield better performance (see~\cref{tab_llm_benchmarking_detail} for full details).
Overcoming this requires genuine reasoning abilities to navigate open-ended action spaces.
}
\centering
\label{tab_rq1_performance}
\begin{tabular}{llcccccc}
\toprule
\textbf{Model} & \textbf{Type} & \textbf{Pass Rate (\%)$^\dagger$} & \makecell[c]{\textbf{Avg.}\\ \textbf{Turns}} & \makecell[c]{\textbf{Avg.}\\ \textbf{Tokens (K)}} & \makecell[c]{\textbf{Avg.}\\ \textbf{Time (min)}} & \textbf{Cost (\$)} & \textbf{\$/Pass} \\
\midrule
Claude Opus 4.7   & Closed & 62.5 (64.8) & 16.7 & 261.5 & 3.6 & 63.47 & 0.51 \\
Gemini 3.1 Pro    & Closed & 55.0 (57.0) & 10.6 & 173.2 & 4.0 & 56.82 & 0.52 \\
Qwen3.6-Max-Preview & Closed & 54.0 (56.8) & 12.5 & 172.2 & 6.9 & 21.44 & 0.20 \\
GPT-5.5           & Closed & 53.5 (56.6) & 14.8 & 499.0 & 3.0 & 100.28 & 0.94 \\
\midrule
Kimi K2.6         & Open   & 57.5 (60.2) & 16.9 & 289.6 & 5.5 & 17.68 & 0.15 \\
GLM 5.1         & Open   & 57.0 (58.5) & 15.5 & 189.1 & 7.0 & 18.24 & 0.16 \\
DeepSeek-V4-Pro     & Open   & 50.0 (52.1) & 20.1 & 398.0 & 9.4 & 17.35 & 0.17 \\
MiniMax M2.7         & Open   & 49.0 (50.8) & 27.5 & 683.6 & 9.3 & 10.95 & 0.11 \\
\bottomrule
\end{tabular}
\vspace{-1em}
\end{minipage}
\end{table*}

\subsection{The \textsc{Verified} Subset}

From the full benchmark, we curate a \textsc{Verified} subset of 200 tasks by carefully balancing diversity and complexity.
It spans diverse real-world task categories in~\cref{fig_stat_a_task_type},
while prioritizing tasks with longer command sequences and non-trivial domain-specific tools.

This subset undergoes manual review by four authors, each with over three years of terminal-based software development experience.
Annotators build the Docker image, enter the containerized environment, and execute the reference solution step by step, repairing the \texttt{Dockerfile} or \texttt{docker-compose.yaml} if necessary.
They also cross-review artifact alignment to ensure that every test constraint is stated in the instruction and that the tests accurately check the persistent artifacts produced by the reference solution.
Tasks that pass this human verification form the highest-fidelity evaluation core of \benchmark.
While the full set of 1,530 tasks provides a scalable resource of in-the-wild terminal workflows, this \textsc{Verified} subset offers a rigorous and challenging testbed for benchmarking frontier models and agents on complex, real-world terminal tasks.
A comparison of the category distribution and task complexity between the full benchmark and \textsc{Verified} is provided in~\cref{appendix_full_verified_comparison}.

\parabf{Evaluation Setup.}
We use \benchmark-\textsc{Verified} to evaluate four frontier closed-source models (\emph{i.e.}, Claude Opus 4.7~\citep{claude_opus_4_7}, GPT-5.5~\citep{gpt_5_5}, Gemini 3.1 Pro~\citep{gemini_3_1_pro}, and Qwen3.6-Max-Preview~\citep{qwen_3_6_max_preview}) alongside four leading open-weight counterparts (\emph{i.e.}, DeepSeek-V4-Pro~\citep{deepseek_v4}, Kimi K2.6~\citep{kimi_k2_6}, GLM-5.1~\citep{glm_5_1}, and MiniMax M2.7~\citep{minimax_m2_7}).
We also assess three general-purpose terminal agents (\emph{i.e.}, Terminus 2~\citep{merrill2026terminalbench}, mini-SWE-agent~\citep{yang2024swe_agent}, OpenHands~\citep{wang2025openhands}) and three model-native CLI assistants (\emph{i.e.}, Claude Code~\citep{claude_code}, Codex CLI~\citep{codex_cli}, Gemini CLI~\citep{gemini_cli}).
All evaluations are conducted using Terminal-Bench's standard Harbor harness~\citep{merrill2026terminalbench}.
Detailed evaluation settings are provided in~\cref{appendix_experiments}.

\begin{figure}[!t]
    \centering
    \includegraphics[width=0.96\linewidth]{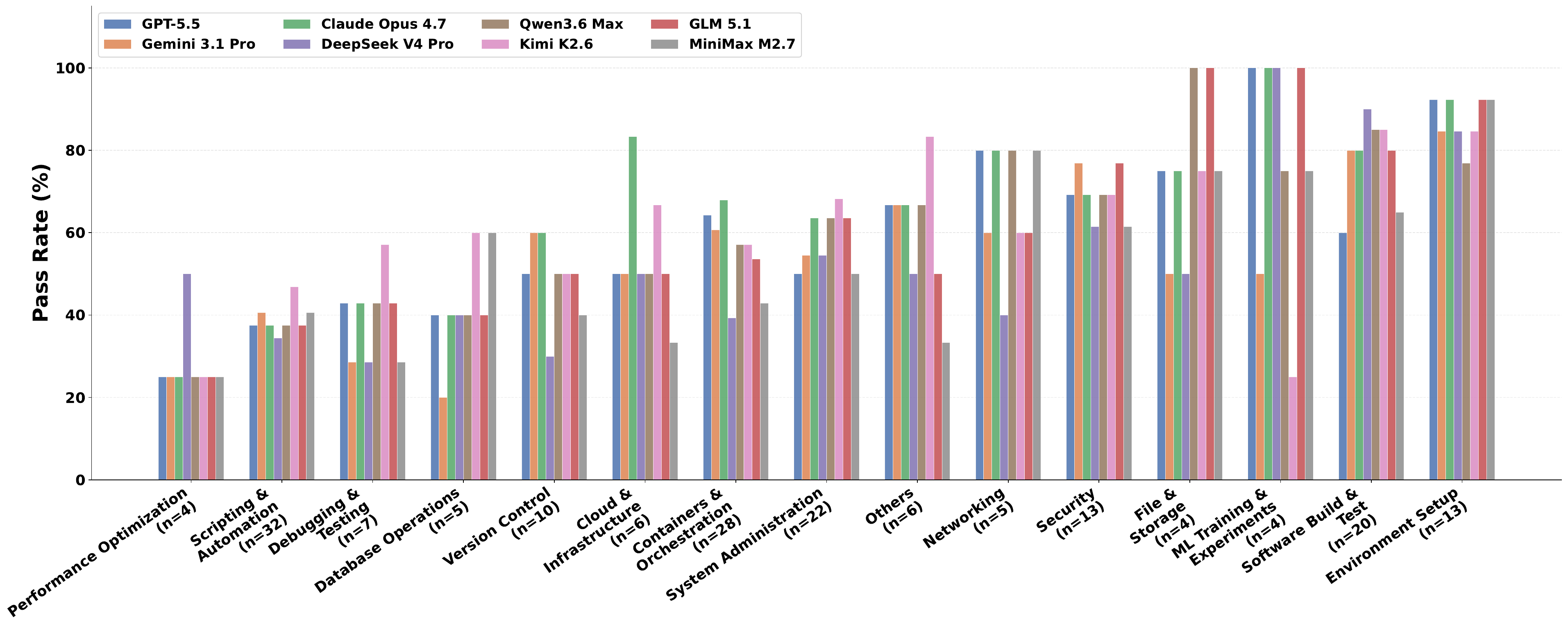}
    \caption{\textbf{Performance of frontier LLMs across terminal task categories.}
    The results indicate that LLMs still struggle with complex tasks involving performance optimization, scripting \& automation, and debugging \& testing, showing domain-specific blind spots and a lack of general tool-use ability.
    Results are computed on the 179 tasks with no evaluation errors across all eight models.}
    \label{fig_rq1_category}
    \vspace{-1em}
\end{figure}

\section{Benchmarking Results on \benchmark-\textsc{Verified}}
\label{sec:experiments}

We evaluate frontier LLMs and terminal agents on \benchmark-\textsc{Verified} to assess model capability and the impact of agent framework choices. We further compare model performance on \benchmark-\textsc{Verified} against Terminal-Bench 2.0~\citep{merrill2026terminalbench} and analyze how agent behavior diverges from the original human workflows.

\subsection{\benchmark-\textsc{Verified} Challenges Large Language Models}

As shown in~\cref{tab_rq1_performance}, we assess frontier LLMs using Terminus-2~\citep{merrill2026terminalbench}, the standardized agent scaffold from Terminal-Bench, to isolate model capability from agent framework differences.

\parabf{Real-World Terminal Tasks Continue to Challenge SOTA LLMs.}
Overall, all evaluated models achieve modest pass rates (49.0\%--62.5\%, avg. 54.8\%), with even the best model (\emph{i.e.}, Claude Opus 4.7) failing on over one-third of the tasks, confirming that the real-world terminal tasks in \benchmark pose a substantial challenge to frontier LLMs.
Interestingly, open-weight models (\emph{e.g.}, Kimi K2.6 and GLM 5.1) demonstrate highly competitive capabilities, rapidly closing the gap with and even outperforming closed-source counterparts (\emph{e.g.}, Gemini 3.1 Pro and GPT-5.5).
Furthermore, they incur significantly lower costs (avg. \$17.13) than closed-source models (avg. \$70.82), yielding a $4\times$--$8\times$ advantage in cost-effectiveness (\$/Pass).

\begin{table*}[!t]
\small
\setlength{\tabcolsep}{3.7pt}
\begin{minipage}[t]{\textwidth}
\caption{\textbf{Performance of terminal agents on \benchmark.} $^\dagger$Values in parentheses denote resolved rate = pass / (total - errors), where errors are tasks on which the Harbor evaluation harness failed before the agent could attempt the task (see~\cref{appendix_agent_benchmarking} for error details).
The results suggest that agent frameworks drive cost-effectiveness rather than shifting the model's capability ceiling (see~\cref{tab_agent_benchmarking_detail} for full details).
Agent designs in real-world environments should prioritize minimizing exploration friction to discover solution paths economically rather than inflating model capabilities.
}
\centering
\label{tab_rq2_framework}
\begin{tabular}{llcccccc}
\toprule
\textbf{Agent} & \textbf{Model} & \textbf{Pass Rate (\%)$^\dagger$} & \makecell[c]{\textbf{Avg.}\\ \textbf{Turns}} & \makecell[c]{\textbf{Avg.}\\ \textbf{Tokens (K)}} & \makecell[c]{\textbf{Avg.}\\ \textbf{Time (min)}} & \textbf{Cost (\$)} & \textbf{\$/Pass} \\
\midrule
Terminus-2  & Claude Opus 4.7 & 62.5 (64.8) & 16.7 & 261.5 & 3.6 & 63.47 & 0.51 \\
Claude Code           & Claude Opus 4.7  & 58.0 (60.7) & 18.0 & 667.7 & 6.0 & 105.12 & 0.91 \\
mini-SWE-agent        & Claude Opus 4.7 & 52.0 (59.8) & 17.1 & 206.4 & 3.9 & 56.94 & 0.55 \\
OpenHands             & Claude Opus 4.7 & 45.0 (57.3) & 21.9 & 410.9 & 5.3 & 371.21 & 4.12 \\
\midrule
Terminus-2    & Gemini 3.1 Pro & 55.0 (57.0) & 10.6 & 173.2 & 4.0 & 56.82 & 0.52 \\
Gemini CLI   & Gemini 3.1 Pro   & 56.0 (59.6) & 41.5 & 694.7 & 6.3 & 85.90 & 0.77 \\
\midrule
Terminus-2           & GPT-5.5 & 53.5 (56.6) & 14.8 & 499.0 & 3.0 & 100.28 & 0.94 \\
Codex CLI    & GPT-5.5          & 48.5 (56.1) & 29.3 & 431.4 & 1.6 & 128.80 & 1.33 \\
\bottomrule
\end{tabular}
\vspace{-1em}
\end{minipage}
\end{table*}

\parabf{LLMs Exhibit an Efficiency Paradox in Real-World Terminal Environments.}
Higher resource consumption does not correlate with better outcomes.
At a statistical level, task success rates show weak negative correlations with both turn count (Pearson $r = -0.49$) and token usage ($r = -0.62$).
This trend is evident in specific models: GPT-5.5 and MiniMax M2.7 consume substantially more tokens and turns than most peers, yet achieve lower pass rates.
Our trajectory analysis further shows that failed attempts are disproportionately expensive.
Across all models, failed attempts consume on average $3.3\times$ more tokens and $1.4\times$ more time than successful ones, monopolizing 63\% of total evaluation costs despite representing only 43\% of attempts.

This \textit{efficiency paradox} reflects a unique challenge of \benchmark: real-world terminal tasks expose open-ended action spaces filled with nuanced dependencies and domain-specific tools.
Without reliable planning and stopping criteria, agents cannot simply brute-force their way to a solution; they may keep exploring the authentic environment, spending more compute without making progress toward the correct outcome.

\parabf{\benchmark Exposes Polarized Tool-Use Capabilities.}
As shown in~\cref{fig_rq1_category}, model performance varies sharply across task domains.
LLMs perform well on environment setup (avg. 87.5\%) and software build \& test (avg. 78.1\%), but struggle with performance optimization (avg. 28.1\%), scripting \& automation (avg. 39.1\%), and debugging \& testing (avg. 39.3\%).
Moreover, no single model dominates across all domains: Claude Opus 4.7 performs strongly on cloud \& infrastructure (83.3\%) and containers \& orchestration (67.9\%), while Kimi K2.6 outperforms it on scripting \& automation (46.9\% vs.\ 37.5\%).
These results show that current LLMs still lack robust tool-use ability across diverse real-world CLI workflows, a blind spot exposed by \benchmark's broad task coverage.

\subsection{Agentic Leaderboarding on \benchmark-\textsc{Verified}}

\cref{tab_rq2_framework} evaluates general terminal agents and model-native CLI assistants across various LLMs.
Note that the Harbor harness~\citep{merrill2026terminalbench} is natively compatible with Terminus-2.
For other agents, the Harbor harness occasionally triggers infrastructural errors (e.g., container initialization timeouts or dependency conflicts) \textit{before} the agent can even attempt the task (see~\cref{appendix_agent_benchmarking} for error details).
Thus, we also report \textit{pass / (total - errors)} in parentheses for fair comparison.

\begin{wrapfigure}{r}{0.45\textwidth}
  \centering
  \vspace{-1em}
  \includegraphics[width=0.45\textwidth]{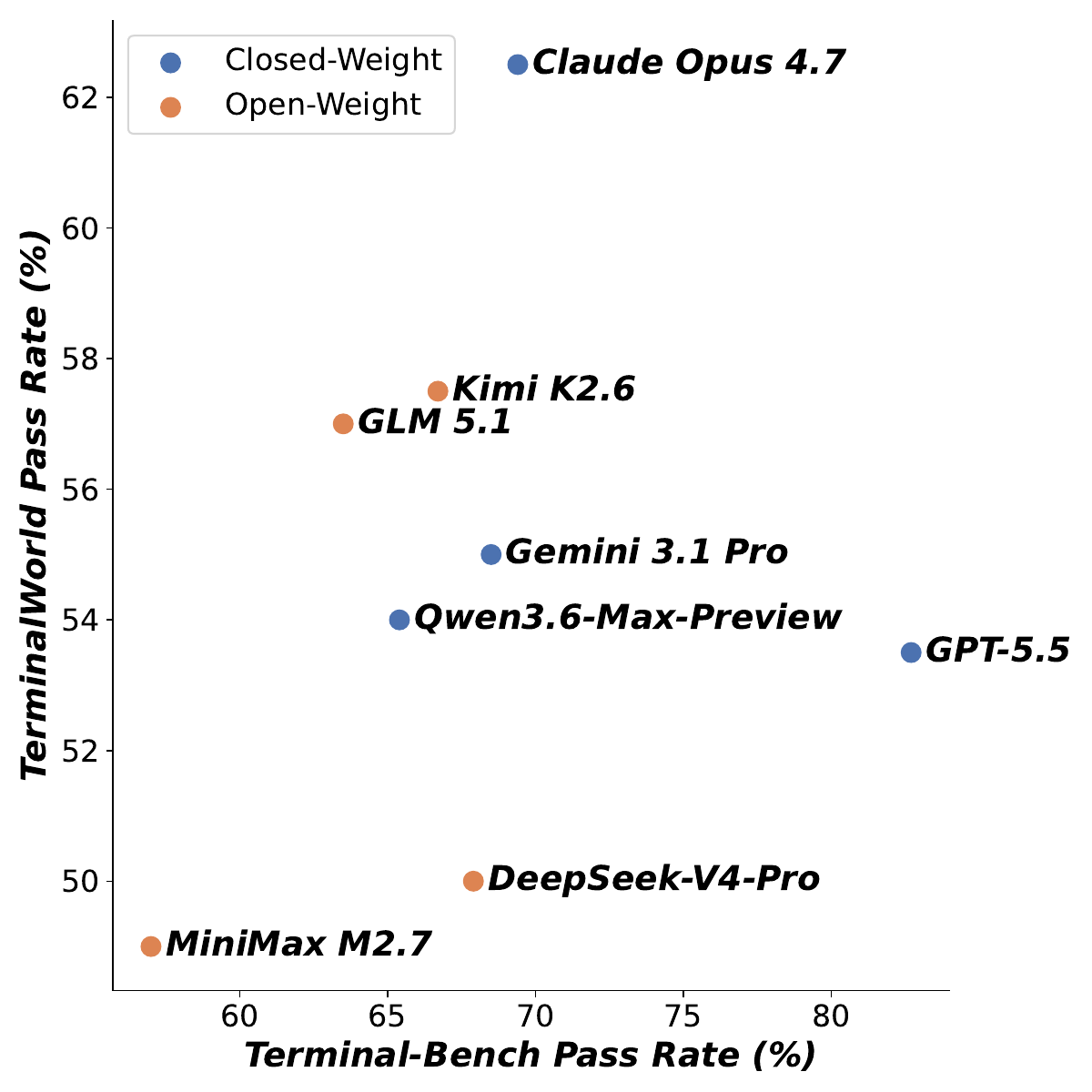}
  \caption{\textbf{Scatter of Terminal-Bench 2.0 score vs.\ \benchmark-\textsc{Verified} score.}
  The weak score correlation (\textbf{Pearson} $\mathbf{r=0.20}$) highlights a disconnect: LLMs that dominate expert-curated challenges still struggle on real-world workflows, indicating that \benchmark assesses a distinct terminal capability.
  }
  \label{fig_rq3_a_score_corr}
  \vspace{-4em}
\end{wrapfigure}

\parabf{Agent Frameworks Drive Cost-Effectiveness More Than Capabilities.}
The results suggest that agent frameworks affect performance, but do not substantially change the underlying model's capability ceiling.
Their larger impact is on \textit{cost-effectiveness}.
For example, with Claude Opus 4.7, pass rates remain relatively close across frameworks, yet Terminus-2 and mini-SWE-agent achieve comparable performance at roughly \$60 total cost (\$0.51--\$0.55 per pass), using fewer turns, tokens, and time.
This suggests that practical agent design for real-world terminal environments should prioritize reducing exploration friction rather than merely adding orchestration complexity, helping models find effective solution paths earlier and more economically.

\subsection{\benchmark vs.\ Terminal-Bench}

To test whether Terminal-Bench performance transfers to real-world terminal workflows, \cref{fig_rq3_a_score_corr} plots each model's Terminal-Bench 2.0 pass rate against its \benchmark-\textsc{Verified} pass rate.
We use the officially reported Terminal-Bench 2.0 scores as each model's reference performance; details are provided in~\cref{appendix_bench_perf_comp}.

\parabf{Terminal-Bench Scores Transfer Weakly to Real-World Performance on \benchmark.}
The results reveal a clear gap between expert-curated challenges and authentic terminal workflows.
As shown in~\cref{fig_rq3_a_score_corr}, \benchmark-\textsc{Verified} presents a harder real-world evaluation setting: models score 57.0\%--82.7\% on Terminal-Bench 2.0, but only 49.0\%--62.5\% on \benchmark-\textsc{Verified}.
Moreover, model rankings are notably reshuffled across the two benchmarks.
GPT-5.5, despite scoring near 83\% on Terminal-Bench 2.0, reaches only 53.5\% on \benchmark-\textsc{Verified}.
In contrast, open-weight models such as Kimi K2.6 achieve moderate Terminal-Bench 2.0 scores (66.7\%) but remain competitive on \benchmark-\textsc{Verified} ($\sim$57\%), outperforming GPT-5.5 and Gemini 3.1 Pro.
Only Claude Opus 4.7 performs strongly across both benchmarks.
This rank reshuffling yields a weak correlation between Terminal-Bench 2.0 and \benchmark-\textsc{Verified} performance (\textbf{Pearson} $\mathbf{r=0.20}$).
This divergence reflects the broader range of terminal capabilities captured by \benchmark, particularly the ability to discover and use domain-specific tools that are underrepresented in Terminal-Bench.
Our analysis shows that, in 55.3\% of failed trajectories, agents never invoke a domain-specific tool used by the reference solution.
For example, on a task requiring a DjVu file, an agent spent 56 episodes manually constructing binary structures without attempting the pre-installed utilities \texttt{cjb2} and \texttt{djvumake}.
Thus, \benchmark captures terminal capabilities beyond Terminal-Bench, showing that expert-curated scores can overestimate competence in authentic terminal environments.

\subsection{Agents vs. Humans}

\begin{wrapfigure}{r}{0.45\textwidth}
  \centering
  \vspace{-1em}
  \includegraphics[width=0.45\textwidth]{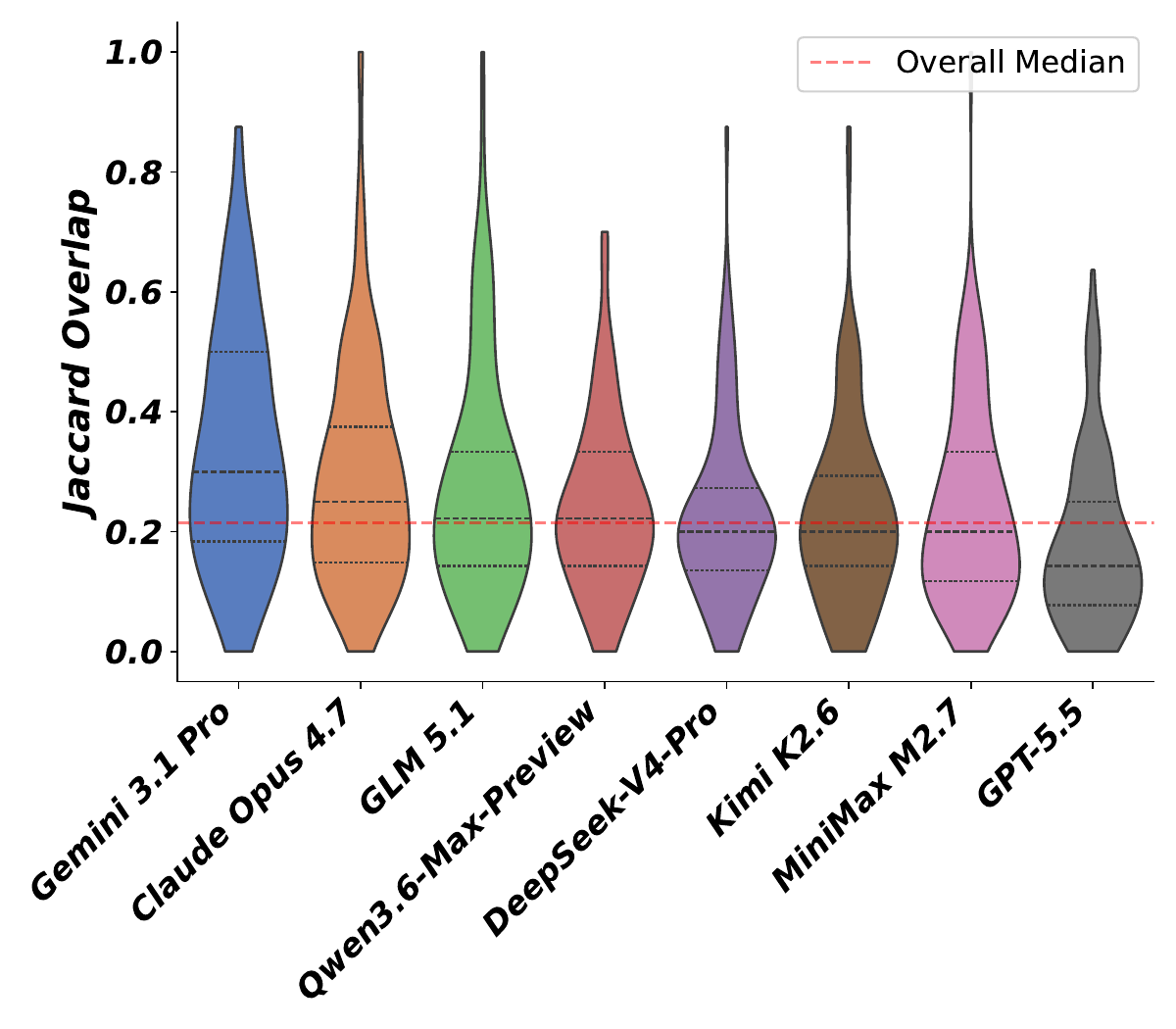}
  \caption{\textbf{Command-Set Overlap of Agents with Human Workflows.}
  While tasks are derived from real-world human recordings, agents often reach the correct outcome through different command paths, with \textbf{21.4\%} median command-set overlap.
  }
  \label{fig_rq4_c_overlap}
  \vspace{-4em}
\end{wrapfigure}

Since each \benchmark task derives from a human recording, we can compare agent behavior against the original human workflow, \textit{e.g.}, whether agents follow similar strategies to humans.
For each successfully solved task, we compute the Jaccard similarity between the command sets in the agent trajectory and the reference solution, after stripping flags and arguments.
\cref{fig_rq4_c_overlap} shows the resulting distribution of command-set overlap across models.
More detailed analysis is provided in~\cref{appendix_agent_human_comp}.

\parabf{Agents Solve Tasks Through Alternative Command Paths, Not Mimicking Humans.}
The median overlap is only \textbf{21.4\%}, meaning agents typically reach the correct outcome via a different set of commands than the human practitioner used.
For example, on a network packet analysis task (extract HTTP Basic Auth credentials from a pcap file), the reference solution uses \texttt{ettercap} to replay and parse the capture; agents instead use \texttt{tshark} with Python to parse the pcap directly.
On a disk image modification task, the reference solution manually creates device nodes with \texttt{mknod} to access partitions; agents use standard tools (\texttt{fdisk}, \texttt{mkfs.ext4}, \texttt{mount}) to achieve the same result.
In both cases, the agent and reference command sets are disjoint, yet the outcome-based verifier accepts both as correct solutions.
This reflects a core design choice of \benchmark: tasks specify the required end state rather than the steps to reach it, so any correct path is accepted.

\section{Conclusion}

We presented \benchmark, a scalable data engine that reverse-engineers high-fidelity terminal evaluation tasks from in-the-wild human recordings.
From 80,870 raw \texttt{asciinema} recordings, we distilled 1,530 validated tasks, including 200 human-verified tasks (\textsc{Verified} subset).
Benchmarking frontier LLMs and agents on this subset shows that current systems still struggle with real-world terminal workflows and that expert-curated benchmark performance does not fully transfer.
As developer practices evolve and new terminal recordings accumulate, \benchmark can be re-run to provide a continually updated testbed for terminal agents.
We hope \benchmark helps the community move beyond static, expert-curated benchmarks toward evaluating and developing agents that are reliable in real-world, evolving terminal workflows.

\section*{Acknowledgments}

We gratefully acknowledge Amazon Web Services (AWS) for supporting this research through AWS credits and Bedrock access support. These resources were used in part for large-scale task synthesis, benchmark construction, and model evaluation in \benchmark. We also thank the AWS team supporting UCL for their guidance and support.

\balance
{
    \small
    \bibliographystyle{unsrtnat}
    \bibliography{references}
}

\newpage
\appendix

\appendix

\section{Broader Impact and Ethical Considerations}
\label[appendix]{sec:ethics}

In developing the \benchmark benchmark, we adhere strictly to ethical data practices and copyright compliance, specifically addressing the challenges inherent in sourcing in-the-wild, user-generated terminal recordings.

\parabf{Data Sourcing and Consent.}
\benchmark sources data from \texttt{asciinema}, a platform where practitioners publicly broadcast terminal sessions for education and collaboration.
Our framework essentially automates this intended use: it ``views'' these recordings to learn from them, just as a human user would.
To ensure ethical compliance at scale, we strictly retrieve publicly listed transcripts using the platform’s standard download mechanisms in full accordance with \texttt{robots.txt} directives.
We collect only the \texttt{.txt} transcripts and their coupled metadata (\emph{e.g.}, titles and environmental specs), which are intentionally exposed by the platform UI and are essential for evaluating task correctness.
All data is utilized strictly for non-commercial academic research.

\parabf{Copyright, Rehosting, and the Right to be Forgotten.}
A critical ethical mandate is respecting the intellectual property of individual creators and the platform's Terms of Service. To definitively prevent the unauthorized redistribution (rehosting) of copyrighted material, \benchmark \emph{does not} distribute the original human transcripts or native \texttt{.cast} files. Instead, our publicly released benchmark tasks contain only the synthesized evaluation artifacts (instructions, reference solutions, execution environments, and test suites) accompanied by direct hyperlinks pointing back to the original recordings on \texttt{asciinema}. This pointer-based architecture ensures robust compliance: it inherently respects the \textit{``Right To Be Forgotten''} of recording authors. If a creator removes their recording from the host platform, our reference hyperlink naturally expires, ensuring no unauthorized local copies persist in our dataset.

\parabf{Privacy and PII Mitigation.}
Terminal recordings inherently carry the risk of exposing sensitive system information or Personally Identifiable Information (PII). As detailed in our data filtering methodology (\cref{sec_data_collection}), we mitigate this risk at the source. Our automated pipeline actively filters out recordings containing exposed PII, sensitive credentials (\emph{e.g.}, API keys, AWS tokens), or operations demonstrating malicious intent before any task synthesis occurs.

\parabf{Human Quality Verification.}
To ensure the integrity of our evaluation core, the final \textsc{Verified} subset of \benchmark tasks underwent rigorous manual verification. This process was conducted entirely by the paper's authors and expert collaborating researchers, ensuring that no uncompensated or under-compensated crowdsourced labor was utilized.

\parabf{Platform Mutual Benefit.}
Rather than exploiting the host platform, \benchmark is designed to create a mutually beneficial ecosystem. By relying on direct hyperlinks rather than rehosted files, our benchmark actively drives community traffic back to \texttt{asciinema}.
Furthermore, this work highlights the broader value of platforms such as \texttt{asciinema} in the LLM era: they preserve authentic human interaction traces that can support the evaluation and improvement of AI agents.
This creates opportunities for platform maintainers and research communities to collaborate on responsible data access, curated recording collections, and benchmark development.

\section{Terminal Task Scope and Categories in \benchmark}
\label[appendix]{appendix_task_category}

\benchmark operationalizes a terminal task as a \textit{pure CLI} workflow: the agent issues shell commands and observes their \texttt{stdout}/\texttt{stderr} output to decide the next action, without invoking any full-screen TUI application.
During data collection (\cref{sec_data_collection}), all recordings involving terminal-based editors (\emph{e.g.}, \texttt{vim}, \texttt{nano}, \texttt{emacs}) or TUI programs are discarded.
Evaluating TUI-based interactions is left to future work.

\parabf{Categories.}
\benchmark covers 18 real-world terminal task categories derived from the purposes of the source recordings.
Each task receives a \emph{single} category label reflecting its \emph{primary goal}, \emph{i.e.}, the final state the practitioner wants to achieve, rather than the incidental tools used along the way.
For example, a task that writes a shell script to automate nightly database backups is categorised as \emph{Database Operations}, not \emph{Scripting \& Automation}, because the script is merely the means (the \emph{journey}), not the end goal.
The full taxonomy with per-category scope boundaries and disambiguation rules is released with the benchmark.
\cref{tab_task_categories} lists all 18 categories with descriptions and task counts.

\begin{table*}[t]
\caption{The 18 task categories in \benchmark, grouped by practitioner goals.
Representative CLI tools are shown in parentheses.
}
\label{tab_task_categories}
\centering\small
\setlength{\tabcolsep}{2pt}
\begin{tabular}{@{}lp{9.4cm}r@{}}
\toprule
\textbf{Category} & \textbf{Description} & \textbf{\#} \\
\midrule
\multicolumn{3}{@{}l}{\textit{Infrastructure \& Operations}} \\[2pt]
System Administration       & Configure, maintain, or repair OS-level infrastructure: services, users, disk partitions, kernel parameters, and boot (\texttt{systemctl}, \texttt{fdisk}, \texttt{sysctl}).  & 191 \\
Networking                  & Configure network services, diagnose connectivity, set up VPN tunnels, or measure network bandwidth (\texttt{dig}, \texttt{wg}, \texttt{haproxy}, \texttt{iperf3}).            & 34 \\
File \& Storage             & Manage, backup, archive, sync, or transfer files as the primary objective (\texttt{rsync}, \texttt{rclone}, \texttt{tar}, \texttt{restic}).                                   & 29 \\
\midrule
\multicolumn{3}{@{}l}{\textit{Software Development}} \\[2pt]
Software Build \& Test      & Compile, test, lint, or package a code project via CLI build toolchains (\texttt{make}, \texttt{cargo}, \texttt{gradle}, \texttt{pytest}).                                    & 212 \\
Version Control             & Manage code history, resolve conflicts, or rewrite repository history as the primary goal (\texttt{git rebase}, \texttt{git bisect}, \texttt{git filter-repo}).                & 70 \\
Environment Setup           & Prepare a development environment: virtual envs, dependency installation, SDK version management, dotfiles (\texttt{conda}, \texttt{nvm}, \texttt{pyenv}).                    & 100 \\
Debugging \& Testing        & Diagnose a functional failure or verify correctness---crash analysis, regression bisection, log triage (\texttt{strace}, \texttt{gdb}, \texttt{valgrind}).                    & 36 \\
Performance Optimization    & Measure, profile, or benchmark execution performance---flame graphs, latency reports, throughput benchmarks (\texttt{perf}, \texttt{hyperfine}, \texttt{py-spy}).             & 14 \\
\midrule
\multicolumn{3}{@{}l}{\textit{Cloud \& Container Stack}} \\[2pt]
Containers \& Orchestration & Build, run, or manage containers and clusters; the container layer (images, pods, registries) is the primary subject (\texttt{docker}, \texttt{kubectl}, \texttt{helm}).    & 184 \\
Cloud \& Infrastructure     & Provision or manage cloud resources via provider CLIs or IaC tools (\texttt{aws}, \texttt{gcloud}, \texttt{az}, \texttt{terraform}).                                          & 26 \\
Deployment \& CI/CD         & Deliver an application to a target environment or configure a CI/CD pipeline (\texttt{gh workflow}, \texttt{argocd}, \texttt{fly deploy}).                                    & 14 \\
\midrule
\multicolumn{3}{@{}l}{\textit{Data Processing \& Automation}} \\[2pt]
Scripting \& Automation     & Write or run scripts, pipelines, or automation routines; includes driving stdin-consuming programs via pipe or heredoc (\texttt{bash}, \texttt{awk}, \texttt{jq}, \texttt{expect}). & 350 \\
Data Analysis               & Extract insights or a transformed dataset from structured data---log analysis, CSV aggregation, statistical summaries (\texttt{pandas}, \texttt{duckdb}, \texttt{csvkit}).    & 39 \\
Database Operations         & Query, administer, migrate, backup, or tune a database; the database itself is the primary subject (\texttt{psql}, \texttt{mongosh}, \texttt{alembic}).                        & 52 \\
ML Training \& Experiments  & Train, fine-tune, evaluate, or run inference on a machine learning model (\texttt{torchrun}, \texttt{huggingface-cli}, \texttt{svm-train}, \texttt{vllm serve}).               & 18 \\
\midrule
\multicolumn{3}{@{}l}{\textit{Specialized Domains}} \\[2pt]
Security                    & Find vulnerabilities, exploit systems, capture CTF flags, reverse-engineer binaries, or audit security posture (\texttt{nmap}, \texttt{gdb}, \texttt{hashcat}, \texttt{radare2}). & 126 \\
Scientific Computing        & Perform domain-specific computation: bioinformatics pipelines, physics simulations, formal proofs, HPC jobs (\texttt{samtools}, \texttt{gmx}, \texttt{sbatch}, \texttt{lean}). & 30 \\
Media Processing            & Transform, convert, or extract content from audio, video, or image files (\texttt{ffmpeg}, \texttt{sox}, \texttt{convert}, \texttt{yt-dlp}).                                  & 5 \\
\midrule
\multicolumn{2}{@{}l}{\textbf{Total}} & \textbf{1{,}530} \\
\bottomrule
\end{tabular}
\end{table*}

\section{Additional Details on Benchmark Construction}
\label[appendix]{appendix_benchmark_construction}

\subsection{Stage-by-Stage Pipeline Analysis}
\label[appendix]{appendix_pipeline_statistics}

In the \benchmark pipeline, our filtering and selection prioritize substantive, multi-step workflows that can be reproduced, executed, and verified.
Below, we report the data retained and the main reasons for exclusion at each stage of the construction pipeline.

\parabf{Data Filtering (80,870 $\rightarrow$ 9,492).}
Approximately 37\% of all raw recordings are removed by a combined triviality and privacy/safety screen, covering recordings with fewer than three commands, seconds-long sessions, or sensitive content.
Recordings involving TUI/GUI interactions, approximately 27\% of the raw corpus, are excluded to retain pure CLI workflows.
LLM value scoring removes another approximately 8\% of raw recordings as aimless exploratory churn, while under 1\% are dropped because their external resources are no longer accessible.

\parabf{Environment Reproduction (9,492 $\rightarrow$ 5,035).}
Each environment is rebuilt approximately 1.7 times on average through the execution-refinement loop.
Over 90\% of the environments that ultimately fail to build are blocked by agent turn/cost budget exhaustion or harness runtime errors.
Classic irreproducibility failures, such as vanished base images or network errors, account for under 2\%.

\parabf{Trial Validation (5,035 $\rightarrow$ 1,530).}
Approximately 11\% of validated tasks require more than one refinement iteration before passing the AllPassing, Nop, and Partial trials.
Among the tasks dropped at this stage, approximately 61\% are adjudicated infeasible (unsolvable or unverifiable), and approximately 37\% hit cost budgets.

\subsection{Intent Fidelity Audit}
\label[appendix]{appendix_intent_audit}

We conducted an independent manual audit of 150 tasks (approximately 10\%) uniformly sampled from the 1,530 automatically generated tasks, to judge whether the instruction, reference solution, and verifier together preserve the core intent of the original recording.
Three of the authors, each with over three years of terminal-based development experience, independently assessed all 150 tasks under the following protocol:
(1) For each task, the annotator first reads only the original recording transcript (with its title and description) and writes a one-sentence hypothesis of the recorded user's intent, before seeing any generated artifact.
(2) The annotator then reviews the instruction, reference solution, and verifier and scores the task on three levels (\emph{faithful}, \emph{drifted}, \emph{unrelated}) based on whether the artifacts deviate from the recording's intent.

\begin{table*}[htbp]
    \centering
    \caption{Human audit of intent fidelity on 150 randomly sampled tasks.}
    \label{tab_intent_audit}
    \begin{tabular}{lrrrr}
        \toprule
        Annotator & Faithful & Drifted & Unrelated & Faithful (\%) \\
        \midrule
        1 & 139 & 11 & 0 & 92.7 \\
        2 & 135 & 14 & 1 & 90.0 \\
        3 & 133 & 15 & 2 & 88.7 \\
        \midrule
        \textbf{Majority vote} & \textbf{136} & \textbf{14} & \textbf{0} & \textbf{90.7} \\
        \bottomrule
    \end{tabular}
\end{table*}

The results in~\cref{tab_intent_audit} show that the vast majority of automatically generated tasks (90.7\%) faithfully preserve the intent of the original recordings, with substantial inter-annotator agreement (Fleiss $\kappa = 0.67$; pairwise raw agreement 93--95\%).
Among the 14 drifted tasks, the main error types are: (1) the recording's primary goal is missing from the task and only a secondary step survives; (2) an intermediate step or side effect of the recording is mistaken as the task goal; (3) the real workflow is mocked, so the task no longer exercises the original behavior.

\subsection{Curation of the \textsc{Verified} Subset}
\label[appendix]{appendix_full_verified_comparison}

We curate the \textsc{Verified} subset by stratifying across task categories while purposively selecting more complex tasks within each category. As a result, it preserves the category distribution of the full \benchmark set while prioritizing tasks with longer command sequences and more specialized tools, serving as a challenging testbed for assessing frontier LLMs and agents.
\Cref{tab_full_verified_distribution} compares the distribution of the full benchmark (F, 1,530 tasks) with the \textsc{Verified} subset (V, 200 tasks), covering category proportion (Prop.), average number of commands in the reference solution (Len.), and average number of unique commands in the reference solution (Uniq.).

\begin{table*}[htbp]
    \centering
    \small
    \setlength{\tabcolsep}{4pt}
    \caption{Comparison of the full benchmark (F, 1,530 tasks) and the \textsc{Verified} subset (V, 200 tasks). Prop. is the category proportion; Len. and Uniq. are the mean number of commands and unique commands in the reference solution, respectively.}
    \label{tab_full_verified_distribution}
    \begin{tabular}{lrrrrrr}
        \toprule
        \multirow{2}{*}{Task category} & \multicolumn{2}{c}{Prop. (\%)} & \multicolumn{2}{c}{Len.} & \multicolumn{2}{c}{Uniq.} \\
        \cmidrule(lr){2-3}\cmidrule(lr){4-5}\cmidrule(lr){6-7}
        & \multicolumn{1}{c}{F} & \multicolumn{1}{c}{V}
        & \multicolumn{1}{c}{F} & \multicolumn{1}{c}{V}
        & \multicolumn{1}{c}{F} & \multicolumn{1}{c}{V} \\
        \midrule
        Scripting \& Automation    & 22.9 & 17.5 &  9.7 & 13.3 & 6.2 &  7.5 \\
        Software Build \& Test    & 13.9 & 11.5 & 14.3 & 16.7 & 7.3 &  8.6 \\
        System Administration     & 12.5 & 15.0 & 15.6 & 18.1 & 8.7 & 10.4 \\
        Containers \& Orchestration & 12.0 & 14.0 & 11.7 & 15.9 & 5.5 & 6.5 \\
        Security                  &  8.2 &  7.5 & 12.3 & 17.1 & 7.3 &  9.8 \\
        Environment Setup         &  6.5 &  7.0 &  8.6 &  8.8 & 5.6 &  5.6 \\
        Version Control           &  4.6 &  5.0 &  9.8 & 14.4 & 4.0 &  5.4 \\
        Networking                &  2.2 &  3.0 & 11.9 & 12.7 & 8.0 &  9.3 \\
        Cloud \& Infrastructure   &  1.7 &  3.0 &  9.9 &  9.0 & 6.7 &  7.0 \\
        Other categories (9)      & 15.5 & 16.5 & 13.0 & 14.1 & 7.2 &  7.4 \\
        \midrule
        Overall                   & 100  & 100  & 12.0 & 14.8 & 6.7 &  7.9 \\
        \bottomrule
    \end{tabular}
\end{table*}

The results show that every category of the full benchmark is represented in \textsc{Verified}, and no category's proportion shifts by more than about 6 percentage points between the two sets.
The complexity skew itself is moderate: on average, reference solutions in \textsc{Verified} contain about 23\% more commands (Len.) and about 18\% more unique commands (Uniq.) than those in the full set.
This selection makes \textsc{Verified} a challenging testbed while retaining coverage across the full benchmark's task categories.

\section{Detailed Experimental Setup and Analysis}
\label[appendix]{appendix_experiments}

\subsection{Benchmarking Large Language Models}
\label[appendix]{appendix_llm_benchmarking}

\begin{table*}[!t]
\footnotesize
\setlength{\tabcolsep}{3.5pt}
\begin{minipage}[t]{\textwidth}
\caption{\textbf{Detailed per-model statistics for the LLM benchmarking experiment under Terminus-2.}
Timeout (TO) is a subset of Fail; Err counts tasks where the harness failed before the agent could run.
For Avg.\ Turns, Avg.\ Tokens, Avg.\ Time, and \$/task, the overall average is shown first, with per-subset averages for passed\,/\,failed tasks in parentheses; averages are computed only over tasks where the agent ran.
$^\dagger$Resolved rate $=$ pass\,/\,(total $-$ errors).}
\centering
\label{tab_llm_benchmarking_detail}
\resizebox{\textwidth}{!}{%
\begin{tabular}{ll r rrrr rrr rr r}
\toprule
& & & \multicolumn{4}{c}{\textbf{Outcomes}} \\
\cmidrule(lr){4-7}
\textbf{Model} & \textbf{Type} & \makecell[c]{\textbf{Pass Rate}\\\textbf{(\%)$^\dagger$}} & \textbf{Pass} & \textbf{Fail} & \textbf{TO} & \textbf{Err} & \makecell[c]{\textbf{Avg.\ Turns}\\\textbf{(Pass\,/\,Fail)}} & \makecell[c]{\textbf{Avg.\ Tokens (K)}\\\textbf{(Pass\,/\,Fail)}} & \makecell[c]{\textbf{Avg.\ Time (min)}\\\textbf{(Pass\,/\,Fail)}} & \textbf{Cost (\$)} & \textbf{\$/Pass} & \makecell[c]{\textbf{\$/task}\\\textbf{(Pass\,/\,Fail)}} \\
\midrule
Claude Opus 4.7     & Closed & 62.5 (64.8) & 125 & 68 &  2 &  7 & 16.7~(12.0\,/\,25.0) & 261.5~(139\,/\,481)  & 3.6~(3.7\,/\,3.6)  & 63.47  & 0.51 & 0.23\,/\,0.39 \\
Gemini 3.1 Pro      & Closed & 55.0 (57.0) & 110 & 83 &  0 &  7 & 10.6~~(9.1\,/\,12.5) & 173.2~~(84\,/\,289)  & 4.0~(3.6\,/\,4.4)  & 56.82  & 0.52 & 0.20\,/\,0.39 \\
Qwen3.6-Max-Preview & Closed & 54.0 (56.8) & 108 & 82 &  3 & 10 & 12.5~(11.6\,/\,13.2) & 172.2~(120\,/\,237)  & 6.9~(5.9\,/\,7.7)  & 21.44  & 0.20 & 0.09\,/\,0.14 \\
GPT-5.5             & Closed & 53.5 (56.6) & 107 & 82 &  3 & 11 & 14.8~~(7.9\,/\,24.1) & 499.0~~(76\,/\,1068) & 3.0~(2.5\,/\,3.6)  & 100.28 & 0.94 & 0.22\,/\,0.93 \\
\midrule
Kimi K2.6           & Open   & 57.5 (60.2) & 115 & 76 &  4 &  9 & 16.9~(13.4\,/\,22.2) & 289.6~(136\,/\,528)  & 5.5~(4.4\,/\,7.3)  & 17.68  & 0.15 & 0.06\,/\,0.15 \\
GLM 5.1             & Open   & 57.0 (58.5) & 114 & 81 &  8 &  5 & 15.5~(15.2\,/\,16.1) & 189.1~(131\,/\,273)  & 7.0~(6.2\,/\,8.1)  & 18.24  & 0.16 & 0.07\,/\,0.13 \\
DeepSeek-V4-Pro     & Open   & 50.0 (52.1) & 100 & 92 & 12 &  8 & 20.1~(19.7\,/\,20.8) & 398.0~(321\,/\,493)  & 9.4~(8.6\,/\,10.4) & 17.35  & 0.17 & 0.08\,/\,0.10 \\
MiniMax M2.7        & Open   & 49.0 (50.8) &  98 & 95 & 13 &  7 & 27.5~(23.6\,/\,31.7) & 683.6~(378\,/\,1006) & 9.3~(7.9\,/\,10.9) & 10.95  & 0.11 & 0.04\,/\,0.08 \\
\bottomrule
\end{tabular}}
\end{minipage}
\end{table*}

All models in this experiment are evaluated using Terminus-2~\citep{merrill2026terminalbench}, Harbor's native agent scaffold, to ensure a fair comparison that isolates model capability from agent framework differences.
For models supporting configurable reasoning effort, we default to the providers' default settings (\emph{e.g.}, ``medium'' for GPT-5.5 and ``high'' for Claude Opus 4.7) to ensure standardized comparison.
All experiments were run on a CPU-based server that executed the Harbor harness and Docker containers locally, while LLM inference was performed through the corresponding provider APIs.
We did not train or fine-tune any models, and no local GPU computation was required.

\parabf{Model Configuration.}
All Terminus-2 trials share a fixed set of agent-level parameters, read from the \texttt{Terminus2.\_\_init\_\_} defaults in the Harbor adapter (\texttt{terminus\_2.py}):

\begin{itemize}[leftmargin=*, label=$\triangleright$]
    \item \textbf{Sampling temperature:} $T = 0.7$ for all models.
    \item \textbf{Reasoning effort / thinking:} \texttt{reasoning\_effort = None} and \texttt{max\_thinking\_tokens = None}, meaning Terminus-2 does not override provider defaults for extended thinking. Each model uses its native behavior.
    \item \textbf{Interleaved thinking:} \texttt{False}. Reasoning content is not retained in chat history and is not sent to the model in subsequent turns.
    \item \textbf{Context summarization:} \texttt{enable\_summarize = True}, with proactive summarization triggered when free tokens drop below \texttt{proactive\_summarization\_threshold = 8000}. Summarization compresses prior conversation history to keep the agent within context limits.
    \item \textbf{Maximum turns:} unbounded (\texttt{max\_turns = None}). The agent runs until it emits a stop signal, hits a timeout, or encounters an unrecoverable error.
    \item \textbf{LLM backend:} LiteLLM (\texttt{llm\_backend = LLMBackend.LITELLM}), which routes all API calls through a unified interface and handles retry logic at the SDK level.
    \item \textbf{Output parser:} \texttt{parser\_name = "json"}. The agent's LLM responses are parsed in JSON format to extract shell commands, file operations, and control signals.
    \item \textbf{Terminal multiplexer:} \texttt{tmux} with pane dimensions $160 \times 40$ characters.
\end{itemize}

\parabf{Harness Configuration.}
All Terminus-2 experiments share the following Harbor settings unless otherwise noted:

\begin{itemize}[leftmargin=*, label=$\triangleright$]
    \item \textbf{Concurrency:} $n = 4$ for all models (to manage API rate limits and Docker resource contention on a single host).
    \item \textbf{Docker lifecycle:} \texttt{environment.force\_build = false} (reuse cached images) and \texttt{environment.delete = true} (prune containers after each trial to prevent resource accumulation across trials).
    \item \textbf{Network isolation:} Harbor automatically modifies each task's \texttt{docker-compose.yaml} to inject \texttt{network\_mode: none} before container startup. This prevents agents from downloading solutions or consulting external services and ensures the task environment is fully self-contained. No Docker network pool exhaustion was observed across any experiment.
\end{itemize}

\parabf{Error Classification.}
A small fraction of task attempts are classified as \emph{errors}, in which the evaluation harness fails before the agent can attempt the task.
Since all models share the same Terminus-2 scaffold, the error sources are identical across models and fall into two categories:

\begin{itemize}[leftmargin=*, label=$\triangleright$]
    \item \textbf{Tmux session initialization failures.}
    Terminus-2 relies on a \texttt{tmux} session to multiplex shell interaction.
    A race condition in its adapter occasionally sends keystrokes before the session is fully initialized, causing the attempt to abort.
    This is a non-deterministic harness bug independent of the task or model.
    \item \textbf{Container startup timeouts.}
    A small number of tasks use resource-intensive Docker images that exceed the harness startup timeout, preventing the agent from entering the environment.
\end{itemize}

These two sources account for all errors observed in this experiment, with error rates ranging from 2.5\% to 5.5\% across models.
We report both the standard pass rate (pass\,/\,total) and the resolved rate (pass\,/\,(total\,$-$\,errors)) to isolate task-solving capability from harness artifacts.
Agent timeouts are \emph{not} classified as errors; an agent that runs until the time limit without producing a correct answer is counted as a failure.

\subsection{Benchmarking Terminal Agents}
\label[appendix]{appendix_agent_benchmarking}

\begin{table*}[!t]
\footnotesize
\setlength{\tabcolsep}{3.5pt}
\begin{minipage}[t]{\textwidth}
\caption{\textbf{Detailed per-agent statistics for the agent benchmarking experiment.}
Timeout (TO) is a subset of Fail; Err counts tasks where the harness failed before the agent could run.
For Avg.\ Turns, Avg.\ Tokens, Avg.\ Time, and \$/task, the overall average is shown first, with per-subset averages for passed\,/\,failed tasks in parentheses; averages are computed only over tasks where the agent ran.
$^\dagger$Resolved rate $=$ pass\,/\,(total $-$ errors).}
\centering
\label{tab_agent_benchmarking_detail}
\resizebox{\textwidth}{!}{%
\begin{tabular}{ll r rrrr rrr rr r}
\toprule
& & & \multicolumn{4}{c}{\textbf{Outcomes}} \\
\cmidrule(lr){4-7}
\textbf{Agent} & \textbf{Model} & \makecell[c]{\textbf{Pass Rate}\\\textbf{(\%)$^\dagger$}} & \textbf{Pass} & \textbf{Fail} & \textbf{TO} & \textbf{Err} & \makecell[c]{\textbf{Avg.\ Turns}\\\textbf{(Pass\,/\,Fail)}} & \makecell[c]{\textbf{Avg.\ Tokens (K)}\\\textbf{(Pass\,/\,Fail)}} & \makecell[c]{\textbf{Avg.\ Time (min)}\\\textbf{(Pass\,/\,Fail)}} & \textbf{Cost (\$)} & \textbf{\$/Pass} & \makecell[c]{\textbf{\$/task}\\\textbf{(Pass\,/\,Fail)}} \\
\midrule
Terminus-2      & Claude Opus 4.7 & 62.5 (64.8) & 125 & 68 &  2 &  7 & 16.7~(12.0\,/\,25.0) & 261.5~~(139\,/\,481)  & 3.6~(3.7\,/\,3.6) & 63.47  & 0.51 & 0.23\,/\,0.39 \\
Claude Code     & Claude Opus 4.7 & 58.0 (60.7) & 116 & 75 &  1 &  9 & 18.0~(18.1\,/\,17.8) & 667.7~~(572\,/\,816)  & 6.0~(5.5\,/\,7.0) & 105.12 & 0.91 & 0.50\,/\,0.63 \\
mini-SWE-agent  & Claude Opus 4.7 & 52.0 (59.8) & 104 & 70 &  1 & 26 & 17.1~(17.2\,/\,16.8) & 206.4~~(191\,/\,229)  & 3.9~(4.0\,/\,3.8) & 56.94  & 0.55 & 0.28\,/\,0.29 \\
OpenHands       & Claude Opus 4.7 & 45.0 (57.3) &  90 & 67 &  1 & 43 & 21.9~(22.7\,/\,20.7) & 410.9~~(417\,/\,403)  & 5.3~(6.1\,/\,4.6) & 371.21 & 4.12 & 2.19\,/\,2.10 \\
\midrule
Terminus-2      & Gemini 3.1 Pro  & 55.0 (57.0) & 110 & 83 &  0 &  7 & 10.6~~(9.1\,/\,12.5) & 173.2~~~(84\,/\,289)  & 4.0~(3.6\,/\,4.4) & 56.82  & 0.52 & 0.20\,/\,0.39 \\
Gemini CLI      & Gemini 3.1 Pro  & 56.0 (59.6) & 112 & 76 &  6 & 12 & 41.5~(40.9\,/\,42.3) & 694.7~~(673\,/\,726)  & 6.3~(4.0\,/\,9.7) & 85.90  & 0.77 & 0.45\,/\,0.47 \\
\midrule
Terminus-2      & GPT-5.5         & 53.5 (56.6) & 107 & 82 &  3 & 11 & 14.8~~(7.9\,/\,24.1) & 499.0~~~(76\,/\,1068) & 3.0~(2.5\,/\,3.6) & 100.28 & 0.94 & 0.22\,/\,0.93 \\
Codex CLI       & GPT-5.5         & 48.5 (56.1) &  97 & 76 &  0 & 27 & 29.3~(32.4\,/\,26.6) & 431.4~~(464\,/\,416)  & 1.6~(1.6\,/\,1.3) & 128.80 & 1.33 & 0.75\,/\,0.73 \\
\bottomrule
\end{tabular}}
\end{minipage}
\end{table*}

\parabf{Evaluation Harness.}
All agents are evaluated using the Harbor harness~\citep{merrill2026terminalbench}, which provisions a fresh Docker container per task, injects the agent, and collects execution traces.
Terminus-2, as Harbor's native agent, is tightly integrated with the harness and requires only a working \texttt{bash} shell inside the container.
All other agents (\emph{e.g.}, Claude Code, Codex CLI, Gemini CLI, OpenHands, Mini-SWE-Agent) are third-party and must be installed \emph{inside} each task container before execution, which requires the harness to download and configure their respective toolchains (Node.js, npm, Python runtimes, etc.).

\parabf{Harness Configuration (CLI Agent Experiments).}
The CLI agent experiments used Harbor with the following settings, shared with the Terminus-2 experiments where not specified:

\begin{itemize}[leftmargin=*, label=$\triangleright$]
    \item \textbf{Concurrency:} $n = 4$ for all CLI agents, to limit Docker resource pressure and API rate-limit risk on a single host.
    \item \textbf{Retry strategy:} previously errored tasks were re-run after fixing agent installation scripts. Results from the retry phase were merged with non-error results from the original run, keeping the most recent result per task.
    \item \textbf{Docker isolation:} identical \texttt{network\_mode: none} injection, \texttt{force\_build: false}, and \texttt{delete: true} as in the LLM benchmarking experiments (\cref{appendix_llm_benchmarking}).
\end{itemize}

\parabf{Error Classification.}
We classify a task attempt as an \emph{error} when the evaluation harness fails before the agent can attempt the task.
Errors are distinct from task failures, in which the agent executes commands but does not satisfy the verifier.
Across our evaluation, errors fall into three categories:

\begin{itemize}[leftmargin=*, label=$\triangleright$]
    \item \textbf{Tmux session initialization failures.}
    Terminus-2 relies on a \texttt{tmux} session to multiplex shell interaction.
    A race condition in its adapter (\texttt{tmux\_session.py}) occasionally sends keystrokes before the session is fully initialized, causing the attempt to abort.
    This affects only Terminus-2 and is independent of the task or model.
    \item \textbf{Container startup timeouts.}
    A small number of tasks use resource-intensive Docker images that exceed the harness startup timeout, preventing any agent from entering the environment.
    \item \textbf{Third-party agent installation failures on incompatible base images.}
    Approximately 10\% of \benchmark tasks use legacy base images (e.g., Ubuntu 16.04/18.04, CentOS 6/7, Debian 9) whose package repositories are end-of-life.
    On these images, \texttt{apt-get update} or \texttt{yum install} fails, preventing the harness from installing the agent's toolchain inside the container.
    As Harbor's native agent, Terminus-2 does not require in-container installation and is therefore unaffected by this failure mode.
    This is the primary driver of the higher error rates observed for third-party agents (4.5--21.5\%) compared to Terminus-2 (3.5--5.5\%).
\end{itemize}

\subsection{\benchmark vs.\ Terminal-Bench}
\label[appendix]{appendix_bench_perf_comp}

\cref{tab_bench_comp_settings} summarizes the scaffold and reasoning effort used for each model on Terminal-Bench and \benchmark.
For Terminal-Bench, we use each model's official self-reported score, which may reflect different scaffolds and reasoning configurations.
For \benchmark, all models are evaluated under a unified Terminus-2 scaffold with default reasoning effort.
Since re-running all models under identical Terminal-Bench conditions is infeasible, we adopt the best available official score as each model's reference performance.

\begin{table}[h]
\centering
\footnotesize
\caption{Evaluation configurations for the Terminal-Bench and \benchmark experiments in~\cref{fig_rq3_a_score_corr}, including the scaffold and reasoning-effort setting used for each model.}
\label{tab_bench_comp_settings}
\setlength{\tabcolsep}{2.5pt}
\begin{tabular}{llccccccc}
\toprule
\multirow{2}{*}{\textbf{Model}} & \multirow{2}{*}{\textbf{Type}} & \multicolumn{3}{c}{\textbf{Terminal-Bench}} & \multicolumn{3}{c}{\textbf{\benchmark}} \\
\cmidrule(lr){3-5} \cmidrule(lr){6-8}
& & \textbf{Scaffold} & \textbf{Reasoning} & \textbf{Pass (\%)} & \textbf{Scaffold} & \textbf{Reasoning} & \textbf{Pass (\%)} \\
\midrule
Claude Opus 4.7     & Closed & Terminus-2           & max    & 69.4 & Terminus-2 & high (default) & 62.5 \\
Gemini 3.1 Pro      & Closed & Terminus-2           & high   & 68.5 & Terminus-2 & high (default) & 55.0 \\
Qwen3.6-Max-Preview & Closed & Terminus-2           & default & 65.4 & Terminus-2 & default & 54.0 \\
GPT-5.5             & Closed & Codex CLI            & xhigh  & 82.7 & Terminus-2 & medium (default) & 53.5 \\
\midrule
Kimi K2.6           & Open   & Terminus-2           & default & 66.7 & Terminus-2 & default & 57.5 \\
GLM 5.1             & Open   & Terminus-2           & default & 63.5 & Terminus-2 & default & 57.0 \\
DeepSeek-V4-Pro     & Open   & Self-Developed Agent & max    & 67.9 & Terminus-2 & high (default) & 50.0 \\
MiniMax M2.7        & Open   & Claude Code          & default & 57.0 & Terminus-2 & default & 49.0 \\
\bottomrule
\end{tabular}
\end{table}

\subsection{Behavior Comparison between Agents and Humans}
\label[appendix]{appendix_agent_human_comp}

This section extends the agent-human comparison from~\cref{sec:experiments}.
Beyond command-set overlap (21.4\% median Jaccard similarity), we further examine how agent success rate varies with task complexity and how agent command counts relate to reference solution length.

\begin{wrapfigure}{r}{0.45\textwidth}
    \centering
    \vspace{-1em}
    \includegraphics[width=0.45\textwidth]{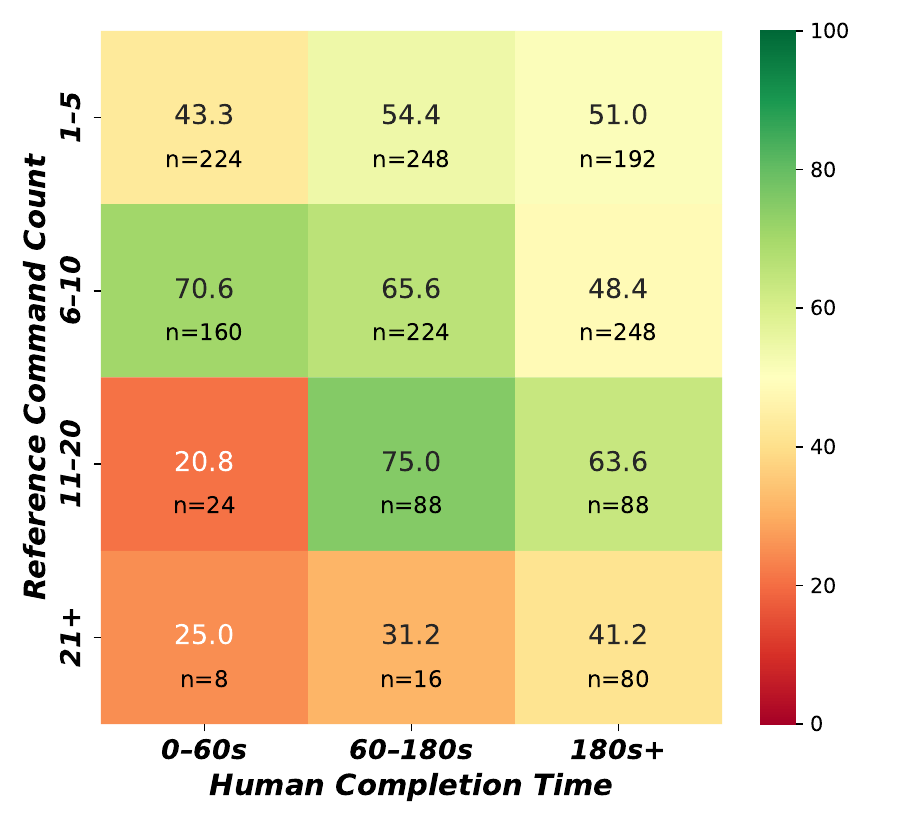}
    \caption{\textbf{Agents vs.\ Humans.}
    Tasks requiring many reference commands are consistently harder for agents, even when humans complete them quickly.}
    \label{fig_rq4_a_heatmap}
    \vspace{-2em}
  \end{wrapfigure}

\parabf{Agent Success Degrades with Procedural Complexity.}
\cref{fig_rq4_a_heatmap} examines pass rate along two dimensions: human completion time and reference command count, where each cell reports the average pass rate across all eight evaluated models on tasks within that bin.
Reference command count is the clearer predictor: tasks requiring 21+ commands remain consistently hard across all time bins (25.0\%--41.2\%), whereas tasks with 6--10 commands achieve up to 70.6\%.
Human completion time, by contrast, is a noisier signal because it conflates two distinct factors: a task with few commands but long runtime (\emph{e.g.}, compilation or model training) appears in the same time bin as a procedurally complex task, yet is fundamentally easier for an agent that only needs to issue the right command and wait.
This is reflected in the data: tasks with 1--5 commands still achieve 51.0\% pass rate even when recordings exceed 180 seconds, while tasks with 21+ commands that humans complete in under 60 seconds drop to 25.0\%.
The key driver of agent difficulty is therefore the length of the sequential command plan the agent must execute correctly, not how long the execution takes.

\begin{figure*}[t]
    \centering
    \includegraphics[width=\textwidth]{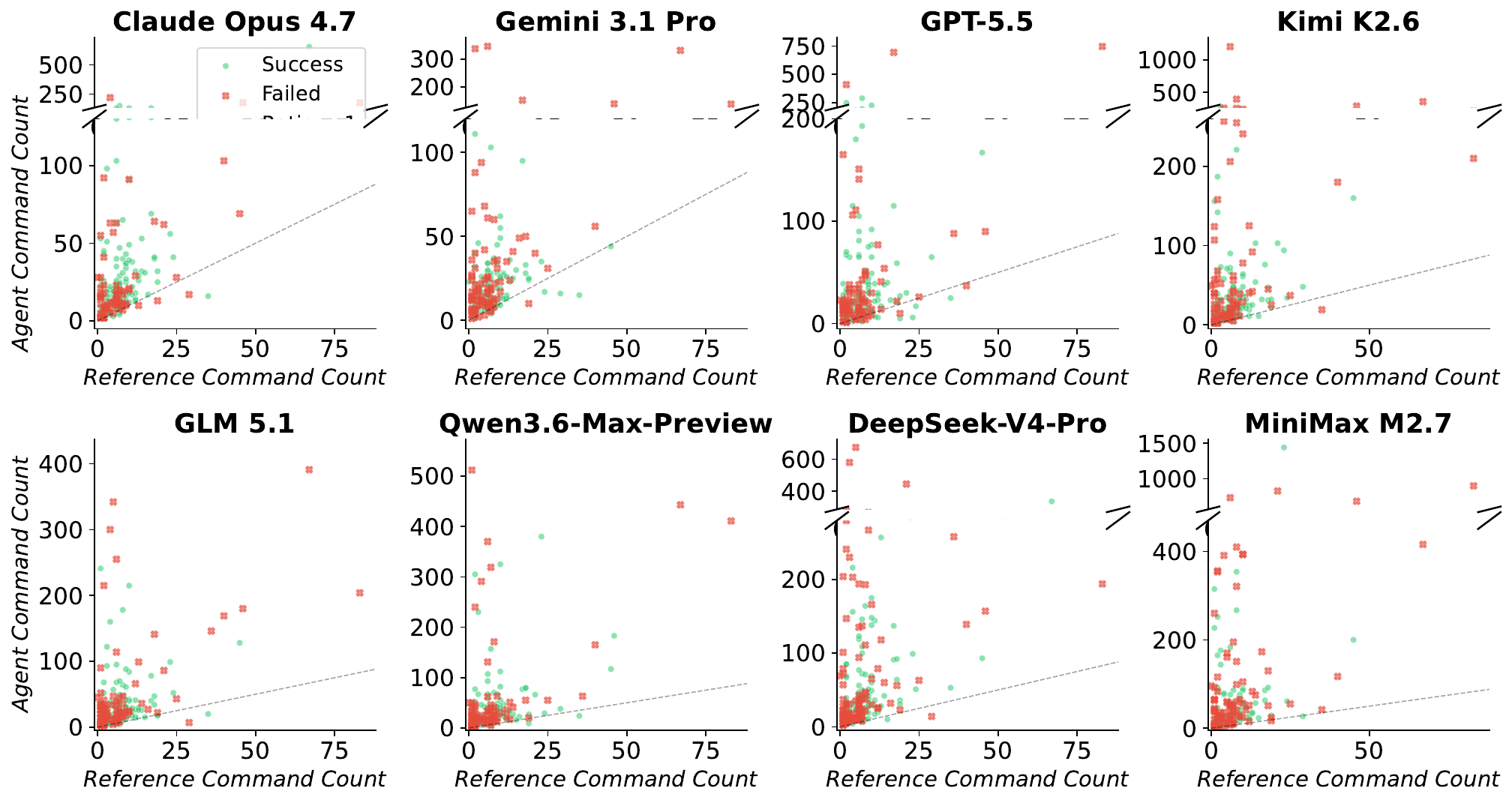}
    \caption{\textbf{Agent Command Count vs.\ Reference Command Count across Eight Models.}
    Reference command count does not tightly predict how many commands an agent will issue. Failed tasks consistently require far more commands than successful ones, reflecting unproductive exploration when the agent cannot identify the correct solution path.}
    \label{fig_rq4_b_grid}
    \vspace{-0.5em}
\end{figure*}

\parabf{Agent Command Count Weakly Follows Reference Workflow Length.}
\cref{fig_rq4_b_grid} compares the agent command count with the reference solution command count across all eight evaluated models.
Although the reference command count captures the length of the original human workflow, it does not tightly predict how many commands an agent will issue and execute.
Most runs, both successful and failed, fall above the diagonal, meaning agents typically issue more commands than the reference solution.
Failed runs cluster further above the diagonal than successful ones, reflecting unproductive exploration when the agent cannot find the correct solution path.
Successful runs are closer to (though still generally above) the diagonal, indicating that agents can reach correct outcomes without needing to match reference solution efficiency exactly.
Overall, agent command count shows little correlation with reference command count across all eight models.
This gap may be explained by the nature of the recordings: source recordings are pre-planned demonstrations rather than exploratory sessions, so reference command counts are compact and clean; agents, by contrast, must explore, verify, and backtrack, accumulating far more commands before converging on a solution.

\parabf{Verifier Validity.}
The trajectories produced by the eight evaluated models also provide evidence about whether the verifiers accept alternative valid solutions and reject plausible but incorrect outcomes.

\begin{itemize}[leftmargin=1.5em, label=$\triangleright$, labelsep=0.5em, itemsep=0.5em, topsep=0.4em, parsep=0pt]
\item \textbf{Acceptance of Alternative Solutions.}
Across successfully solved tasks, the median command-set Jaccard overlap between agent trajectories and reference solutions is only 21.4\%, showing that agents commonly pass the verifiers through command paths that differ substantially from the reference.
For example, on a network packet analysis task (extracting HTTP Basic Auth credentials from a \texttt{pcap} file), the reference solution uses \texttt{ettercap} to replay and parse the capture, while agents instead parse the \texttt{pcap} directly with \texttt{tshark} and Python.
Similarly, on a disk image modification task, the reference solution manually creates device nodes with \texttt{mknod}, while agents achieve the same result with standard tools (\texttt{fdisk}, \texttt{mkfs.ext4}, \texttt{mount}).
In both cases, the agent and the reference share none of the task-critical commands, yet the outcome-based verifier accepts both as correct.

\item \textbf{Rejection of Incorrect Outcomes.}
In 66.4\% of failed trajectories, the agent executed a reasonable workflow yet produced a final state that deviated from the required outcome in a functionally meaningful detail, and the verifier caught the deviation.
For example, on a Kubernetes network policy task, the agent completed an otherwise reasonable configuration workflow but wrote the policy selectors in Kubernetes-native syntax where the instruction explicitly required the Calico-specific syntax; the two are semantically close yet functionally different, and the verifier rejected the resulting policy.
Similarly, on a build task, the agent invoked \texttt{make -f Makefile-v5}, treating the suffix as an alternate makefile, whereas the required invocation was the build target \texttt{make build-v5}; the resulting final state differed from the required one, and the verifier rejected it.
In both cases, the agent's workflow was plausible and close to correct, yet the verifier rejected the final state on precisely the semantic criterion that separates a fully correct outcome from a near miss.
\end{itemize}

Together, these cases show that the verifiers judge whether the required end state is reached, rather than whether the reference commands were reproduced.

\newpage
\section*{NeurIPS Paper Checklist}

\begin{enumerate}

\item {\bf Claims}
    \item[] Question: Do the main claims made in the abstract and introduction accurately reflect the paper's contributions and scope?
    \item[] Answer: \answerYes{}
    \item[] Justification: The abstract and introduction clearly state the paper's main contributions: a scalable data engine for reverse-engineering terminal tasks from in-the-wild recordings, the resulting benchmark and \textsc{Verified} subset, and the main benchmarking findings. These claims are supported by the methodology, dataset statistics, and experiments reported in Sections~3--5.

\item {\bf Limitations}
    \item[] Question: Does the paper discuss the limitations of the work performed by the authors?
    \item[] Answer: \answerYes{}
    \item[] Justification: The paper discusses limitations and risks related to sourcing public terminal recordings, including privacy, copyright, rehosting, and right-to-be-forgotten considerations in~\cref{sec:ethics}. It also reports evaluation limitations such as harness errors, retry handling, and resolved-rate reporting in~\cref{appendix_experiments}.

\item {\bf Theory assumptions and proofs}
    \item[] Question: For each theoretical result, does the paper provide the full set of assumptions and a complete (and correct) proof?
    \item[] Answer: \answerNA{}
    \item[] Justification: The paper does not present theoretical results, formal theorems, or proofs; it focuses on benchmark construction and empirical evaluation.

    \item {\bf Experimental result reproducibility}
    \item[] Question: Does the paper fully disclose all the information needed to reproduce the main experimental results of the paper to the extent that it affects the main claims and/or conclusions of the paper (regardless of whether the code and data are provided or not)?
    \item[] Answer: \answerYes{}
    \item[] Justification: The paper describes the benchmark construction pipeline, evaluation setup, model and agent configurations, Harbor harness settings, retry/error handling, and cost/token accounting in Sections~3--5 and~\cref{appendix_experiments}. We also provide the data and code through the repository linked in the paper, enabling reproduction or verification of the main experimental results subject to access to the evaluated models.

\item {\bf Open access to data and code}
    \item[] Question: Does the paper provide open access to the data and code, with sufficient instructions to faithfully reproduce the main experimental results, as described in supplemental material?
    \item[] Answer: \answerYes{}
    \item[] Justification: Data and code are publicly available through the repository linked in the paper, with the benchmark data hosted on HuggingFace and referenced from the repository. The repository includes instructions for accessing the data, preparing the benchmark tasks, and running the evaluation scripts needed to reproduce the main experimental results, subject to access to the evaluated models.

\item {\bf Experimental setting/details}
    \item[] Question: Does the paper specify all the training and test details (e.g., data splits, hyperparameters, how they were chosen, type of optimizer) necessary to understand the results?
    \item[] Answer: \answerYes{}
    \item[] Justification: The paper specifies the evaluation setup, evaluated models and agents, scaffold and harness configuration, timeout and concurrency settings, retry/error handling, and cost/token accounting in Section~4.2 and~\cref{appendix_experiments}. The work does not involve model training, so optimizer and training-hyperparameter details are not applicable.

\item {\bf Experiment statistical significance}
    \item[] Question: Does the paper report error bars suitably and correctly defined or other appropriate information about the statistical significance of the experiments?
    \item[] Answer: \answerYes{}
    \item[] Justification: The paper reports correlation analyses, including Pearson correlation coefficients for the relationship between Terminal-Bench and \benchmark performance, as well as correlations between success and resource usage. For benchmark pass rates, results are reported descriptively over the fixed \textsc{Verified} task set with task counts and resolved rates; no model training, random splits, or multi-seed training runs are involved.

\item {\bf Experiments compute resources}
    \item[] Question: For each experiment, does the paper provide sufficient information on the computer resources (type of compute workers, memory, time of execution) needed to reproduce the experiments?
    \item[] Answer: \answerYes{}
    \item[] Justification: The paper states that experiments were run on a CPU-based server executing the Harbor harness and Docker containers locally, with LLM inference performed through provider APIs and no local GPU training or fine-tuning. It also reports timeout, concurrency, Docker lifecycle settings, and per-model wall-clock time, token usage, and API cost in Section~5 and~\cref{appendix_experiments}.

\item {\bf Code of ethics}
    \item[] Question: Does the research conducted in the paper conform, in every respect, with the NeurIPS Code of Ethics \url{https://neurips.cc/public/EthicsGuidelines}?
    \item[] Answer: \answerYes{}
    \item[] Justification: We have reviewed the NeurIPS Code of Ethics and believe the research conforms to it. The paper discusses ethical data sourcing, privacy and PII mitigation, copyright and rehosting constraints, right-to-be-forgotten considerations, and responsible release practices in Appendix~A.

\item {\bf Broader impacts}
    \item[] Question: Does the paper discuss both potential positive societal impacts and negative societal impacts of the work performed?
    \item[] Answer: \answerYes{}
    \item[] Justification: Appendix~A discusses broader impacts and ethical considerations, including positive impacts such as improving evaluation of terminal agents and supporting public knowledge-sharing platforms. It also discusses potential risks related to privacy, credentials, copyright, rehosting, and misuse, together with mitigation strategies such as PII filtering, excluding malicious recordings, and avoiding redistribution of original transcripts or cast files.

\item {\bf Safeguards}
    \item[] Question: Does the paper describe safeguards that have been put in place for responsible release of data or models that have a high risk for misuse (e.g., pre-trained language models, image generators, or scraped datasets)?
    \item[] Answer: \answerYes{}
    \item[] Justification: The paper describes safeguards for responsible dataset release in Appendix~A. We filter recordings containing PII, sensitive credentials, or malicious operations; we do not redistribute original transcripts or native cast files; and the released benchmark contains synthesized artifacts with links to the original recordings, supporting removal if the source recording is deleted.

\item {\bf Licenses for existing assets}
    \item[] Question: Are the creators or original owners of assets (e.g., code, data, models), used in the paper, properly credited and are the license and terms of use explicitly mentioned and properly respected?
    \item[] Answer: \answerYes{}
    \item[] Justification: The paper credits the existing assets used in the study, including \texttt{asciinema}, Terminal-Bench/Harbor, agent frameworks, and model providers. Appendix~A describes how we respect the source platform's terms and creators' rights by using publicly available transcripts, following standard access mechanisms, avoiding redistribution of original transcripts or cast files, and releasing only synthesized benchmark artifacts with links to the original recordings.

\item {\bf New assets}
    \item[] Question: Are new assets introduced in the paper well documented and is the documentation provided alongside the assets?
    \item[] Answer: \answerYes{}
    \item[] Justification: The paper introduces the \benchmark benchmark and its \textsc{Verified} subset as new assets. Documentation is provided in the paper and alongside the released repository/dataset, including task structure, benchmark artifacts, data access instructions, evaluation scripts, licensing information, and ethical considerations for data sourcing and release.

\item {\bf Crowdsourcing and research with human subjects}
    \item[] Question: For crowdsourcing experiments and research with human subjects, does the paper include the full text of instructions given to participants and screenshots, if applicable, as well as details about compensation (if any)?
    \item[] Answer: \answerNA{}
    \item[] Justification: The paper does not involve crowdsourcing experiments or direct research with human subjects. The benchmark is derived from publicly shared terminal recordings, and manual verification is conducted by the authors rather than external crowd workers.

\item {\bf Institutional review board (IRB) approvals or equivalent for research with human subjects}
    \item[] Question: Does the paper describe potential risks incurred by study participants, whether such risks were disclosed to the subjects, and whether Institutional Review Board (IRB) approvals (or an equivalent approval/review based on the requirements of your country or institution) were obtained?
    \item[] Answer: \answerNA{}
    \item[] Justification: The paper does not involve crowdsourcing or direct research with human subjects, such as recruiting participants, conducting interventions, or collecting private participant data. The work uses publicly shared terminal recordings and discusses privacy, PII mitigation, and responsible data release in Appendix~A.

\item {\bf Declaration of LLM usage}
    \item[] Question: Does the paper describe the usage of LLMs if it is an important, original, or non-standard component of the core methods in this research? Note that if the LLM is used only for writing, editing, or formatting purposes and does \emph{not} impact the core methodology, scientific rigor, or originality of the research, declaration is not required.
    \item[] Answer: \answerYes{}
    \item[] Justification: LLMs and LLM agents are core components of the proposed data engine, including task instruction formalization, reference solution extraction, environment reproduction, and test-suite generation/refinement. Their roles are described in Section~3.

\end{enumerate}

\end{document}